\documentclass[runningheads,a4paper,12pt]{llncs} 

\usepackage{graphicx}
\usepackage{xcolor}
\usepackage{amsmath}
\usepackage{amssymb}

\usepackage[hidelinks, pdfcreator={}, pdfproducer={}]{hyperref}

\usepackage[left=25mm,right=25mm,top=25mm,bottom=25mm]{geometry}

\usepackage{siunitx}	
\usepackage{subfigure}

\usepackage{tikz}
\usepackage{pgfplots}

\usepackage{enumitem}

\usepackage{acronym}

\tikzset{>=stealth}

\acrodef{1d}[1D]{one dimensional}
\acrodef{2d}[2D]{two dimensional}
\acrodef{3d}[3D]{three dimensional}
\acrodef{4d}[4D]{four dimensional}
\acrodef{2pp}[2PP]{two plane parametrization}
\acrodef{awgn}[AWGN]{additive white Gaussian noise}
\acrodef{cnn}[CNN]{convolutional neural network}
\acrodef{cpu}[CPU]{central processing unit}
\acrodef{l2pp}[L2PP]{local two-plane parametrization}
\acrodef{dof}[DOF]{depth of field}
\acrodef{dslr}[DSLR]{digital single-lens reflex}
\acrodef{dso}[DSO]{Direct Sparse Odometry}
\acrodef{dft}[DFT]{Discrete Fourier Transform}
\acrodef{dslr}[DSLR]{digital single-lens reflex}
\acrodef{epi}[EPI]{epipolar plane image}
\acrodef{egi}[EGI]{extended Gaussian image}
\acrodef{ekf}[EKF]{extended Kalman filter}
\acrodef{fov}[FOV]{field of view}
\acrodef{fir}[FIR]{finite impulse response}
\acrodef{gps}[GPS]{Global Positioning System}
\acrodef{gpu}[GPU]{graphics processing unit}
\acrodef{imu}[IMU]{inertial measurement unit}
\acrodef{icp}[ICP]{Iterative Closest Point}
\acrodef{lrf}[LRF]{laser rangefinder}
\acrodef{logscale}[log-scale]{logarithmized scale}
\acrodef{mae}[MAE]{maximum absolute error}
\acrodef{mla}[MLA]{micro lens array}
\acrodef{mvg}[MVG]{multiple view geometry}
\acrodef{pca}[PCA]{principal component analysis}
\acrodef{rgb}[RGB]{red, green, blue}
\acrodef{rgbd}[RGB-D]{red, green, blue - depth}
\acrodef{rms}[RMS]{root mean square}
\acrodef{rmse}[RMSE]{root mean square error}
\acrodef{sgm}[SGM]{Semi-Global Matching}
\acrodef{snr}[SNR]{signal to noise ratio}
\acrodef{slam}[SLAM]{simultaneous localization and mapping}
\acrodef{sfm}[SfM]{structure from motion}
\acrodef{ssid}[SSID]{sum of squared intensity differences}
\acrodef{tof}[TOF]{time-of-flight}
\acrodef{tcp}[TCP]{total covering plane}
\acrodef{uav}[UAV]{unmanned aerial vehicle}
\acrodef{vo}[VO]{visual odometry}

\definecolor{dgreen}{rgb}{0,0.7,0}				
\colorlet{gColor1}{cyan} 
\colorlet{gColor2}{red} 
\colorlet{gColor3}{dgreen} 
\colorlet{gColor5}{purple} 
\colorlet{gColor8}{brown}

\colorlet{highlightBlue}{blue}

\colorlet{drawRed}{red}
\colorlet{drawBlue}{cyan}
\colorlet{drawGrey}{black!30}

\DeclareMathOperator*{\argmin}{arg~min}

\makeatletter
\renewcommand\subsubsection{\@startsection{subsubsection}{3}{\z@}%
                       {-18\p@ \@plus -4\p@ \@minus -4\p@}%
                       {4\p@ \@plus 2\p@ \@minus 2\p@}%
                       {\normalfont\normalsize\bfseries\boldmath
                        \rightskip=\z@ \@plus 8em\pretolerance=10000 }}
\makeatother

\begin{document}

\title{A Synchronized Stereo and Plenoptic Visual Odometry Dataset}

\author{Niclas Zeller\inst{1,2} \and Franz Quint\inst{2} \and Uwe Stilla\inst{1}}

\institute{
      Technische Universit{\"a}t M{\"u}nchen\\
      \email{niclas.zeller@tum.de}, \email{stilla@tum.de}\and
      Karlsruhe University of Applied Sciences\\
      \email{franz.quint@hs-karlsruhe.de}
}


\maketitle

\begin{abstract}
We present a new dataset to evaluate monocular, stereo, and plenoptic camera based visual odometry algorithms.
The dataset comprises a set of synchronized image sequences recorded by a \ac*{mla} based plenoptic camera and a stereo camera system.
For this, the stereo cameras and the plenoptic camera were assembled on a common hand-held platform.
All sequences are recorded in a very large loop, where beginning and end show the same scene.
Therefore, the tracking accuracy of a visual odometry algorithm can be measured from the drift between beginning and end of the sequence.
For both, the plenoptic camera and the stereo system, we supply full intrinsic camera models, as well as vignetting data.
The dataset consists of 11 sequences which were recorded in challenging indoor and outdoor scenarios.
We present, by way of example, the results achieved by state-of-the-art algorithms.
\end{abstract}

\section{Introduction}
Simultaneous localization and mapping\acused{slam} (\acs{slam}) as well as \ac{vo} based on monocular \cite{Klein2007,Newcombe2011,Engel2013,Forster2014,Engel2014,Mur-Artal2015,Engel2017}, stereo \cite{Engel2015,Mur-Artal2017,Wang2017}, and RGB-D \cite{Izadi2011,Kerl2013,Kerl2015,Mur-Artal2017} cameras have been studied extensively over the last years.
Recently, it was shown that \ac{vo} can also be performed reliably based on plenoptic cameras (or light field cameras) \cite{Dansereau2011,Dong2013,Zeller2017,Zeller2018a}.

While there are various datasets available for traditional types of cameras -- monocular \cite{Zhang2016b,Engel2016a,Majdik2017}, stereo \cite{Geiger2012,Burri2016,Schoeps2017}, and RGB-D \cite{Sturm2012,Handa2014} -- there are no public datasets available for plenoptic camera based \ac{vo}.
The few existing algorithms were only tested on very simple and short sequences.

In this paper, we present a versatile and challenging dataset which supplies synchronizes light field and stereo images.
These sequences were recorded based on a hand-held platform on which a \ac{mla} based plenoptic camera and a stereo camera system is mounted. 

Of course, there exist larger dataset for monocular and stereo cameras.
However, the goal of this dataset is to evaluate the versatility of plenoptic \ac{vo} algorithms and rank these algorithms with respect to methods based on traditional cameras (i.e. monocular and stereo).

The entire synchronized stereo and plenoptic \ac{vo} dataset is available at:
\begin{center}
\textcolor{red}{\url{https://www.hs-karlsruhe.de/odometry-data/}}
\end{center}

\section{Outline}

In Section~\ref{sec:data_acquisition_setup} we present the platform which we used to record the image sequences.
Afterwards, in Section~\ref{sec:stereo_plenoptic_vo_dataset}, we describe the entire structure of the dataset.
This includes the calibration of all cameras as well as the ground truth data and the suggested evaluation metrics.
Section~\ref{sec:results} shows the results of existing algorithms, which were obtained for the proposed dataset.
Furthermore, some 3D reconstructions calculated by \cite{Zeller2018a} are shown to get an impression of the recorded sequences.
Section~\ref{sec:limitations} mentions some limitations, which should be taken into account, when rating the results of different algorithms against each other.


\section{Data Acquisition Setup}\label{sec:data_acquisition_setup}
\begin{figure}[t]
\centering
\includegraphics[width=0.5\linewidth]{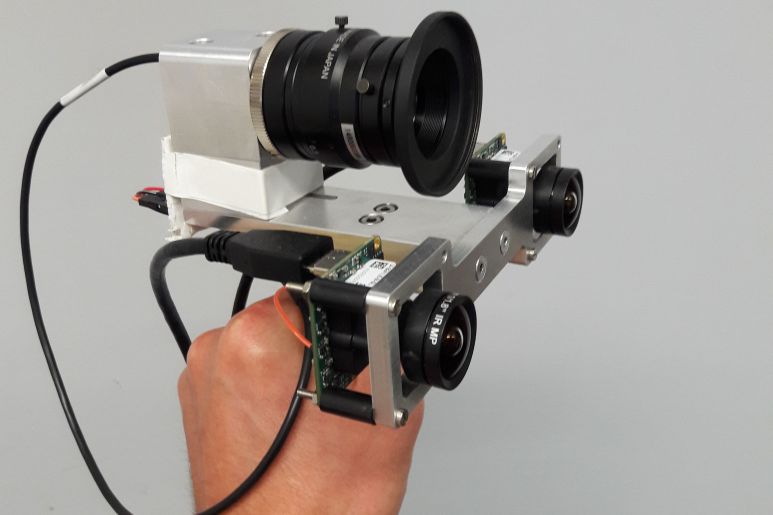}
\caption[Hand-held platform to acquire time synchronized image sequences from a focused plenoptic camera and a stereo camera system]{Handheld platform to acquire time synchronized image streams from a focused plenoptic camera and a stereo camera system.}
\label{fig:stereo_plenoptic_platform}
\end{figure}

To be able to compare plenoptic with stereo or monocular algorithms, a hand-held platform was developed.
On this platform a \ac{mla} based plenoptic camera and a stereo camera system are assembled.
This platform is shown in Figure~\ref{fig:stereo_plenoptic_platform}.

The stereo camera system is based on two monochrome, global shutter, industrial grade cameras by IDS Imaging Development Systems GmbH (model: UI-3241LE-M-GL).
On both cameras, a lens from Lensation GmbH (model: BM4018S118) with \SI{4}{mm} focal length is mounted.
Furthermore, the stereo system has a baseline distance of $\SI{100}{mm}$.

The utilized plenoptic camera is a R5 by Raytrix GmbH.
This camera is based on a xiQ sensor from Ximea GmbH (model: MG042CG-CM-TG).
The xiQ is also a global shutter, industrial grade sensor.
To achieve a suitable trade-off between a wide \ac{fov} and a high angular resolution of the captured light field, a main lens with focal length $f_L = \SI{16}{mm}$ from Kowa (model: LM16HC) was mounted on the camera.

Table~\ref{tab:camera_specifications} lists all important specifications for the plenoptic camera and the stereo camera system.
\begin{table}
\begin{center}
\begin{tabular}{l|rr}
& plenoptic camera & \qquad stereo system\\
\hline
\hline
cameras & 1 & 2\\
pixel size & \SI{5.5}{\micro \meter} & \SI{5.3}{\micro \meter}\\
resolution & \SI{2048 x 2048}{} & \SI{1280 x 1024}{}\\
color channels & 3 & 1\\
focal length & \SI{16}{mm} & \SI{4}{mm}\\
aperture & f/2.8 & f/1.8\\
stereo baseline & -- & \SI{100}{mm}\\
\end{tabular}
\end{center}
\caption[Camera specifications for the plenoptic camera and the stereo camera system]{Camera specifications for the plenoptic camera and the stereo camera system.}
\label{tab:camera_specifications}
\end{table}

To receive synchronized image sequences from all three cameras, one camera of the stereo systems runs in master mode and generates a signal which, in turn, triggers the other two cameras (second camera of the stereo system and plenoptic camera) running in slave mode.
To record the data, all three cameras are connected to a single laptop.
Using this platform, we are able to record synchronized image sequences for all three cameras running at the maximum image resolution and \SI{8}{bit} quantization with frame rates of more than \SI{30}{fps}.

Because of the small \ac{fov}, the images of the plenoptic camera in particular are affected by motion blur.
To keep the motion blur in an acceptable range, for all cameras the exposure time is upper bounded at $\SI{8}{ms}$.
Below this boundary, the automatic exposure adjustment of the respective camera controls the exposure time\footnote{For the stereo cameras, automatic exposure adjustment is only performed for the master (left) camera, while the slave (right) camera adopts the master setting.}.

\section{The Dataset}\label{sec:stereo_plenoptic_vo_dataset}

Based on the platform presented in Section~\ref{sec:data_acquisition_setup}, we recorded a synchronized dataset for the quantitative comparison of plenoptic and stereo \ac{vo} systems.
The inspiration for this dataset was taken from the Benchmark for monocular \ac{vo} presented by Engel et al. \cite{Engel2016a}.

Especially for long, large scale trajectories, it is impossible to obtain reference measurements which are accurate enough to serve as ground truth.
Hence, guided by the idea of \cite{Engel2016a}, we perform all sequences in the dataset as a single very large loop, for which beginning and end of the sequence capture the same scene.

For all recorded sequences we supply geometric camera parameters and vignetting data for both the stereo camera system and the plenoptic camera.
Furthermore, for each sequence, ground truth data is obtained from the loop closure between beginning and end.

\subsection{Geometric Camera Calibration}\label{sec:geometric_camera_calibration}

For all cameras, we perform geometric calibration based on a 3D target, which is presented below in Section~\ref{sec:calibration_approach}.
Aside from the camera models for the plenoptic camera and the stereo system, we supply the entire sets of images as well as the marker positions detected in the images, which were used for calibration.
This way, one can later test new camera models and calibration approaches on the basis of this data.

\subsubsection{Plenoptic Camera Model}

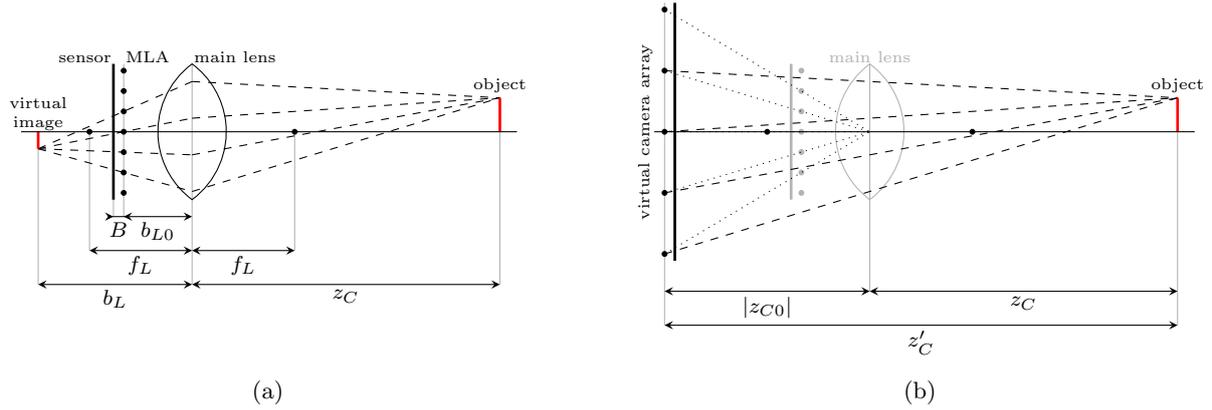
\begin{figure}[t]
\centering
\subfigure[]{

\newcommand*\LensBig[4]{%
  \draw[color=#4] (#1,#2) ++(0,#3/2) arc(180-53.13010235:180+53.13010235:#3*0.625) arc(-53.13010235:53.13010235:#3*0.625) -- cycle ++(0,-#3)
	}

\begin{tikzpicture}[
	scale=0.045,
    >=stealth,
    inner sep=0pt, outer sep=2pt,%
]

\useasboundingbox (-55,-70) rectangle (100,38);

\draw[color = black!30] (0,-45) -- (0,20);
\draw[color = black!30] (-20,-25) -- (-20,20);
\draw[color = black!30] (-23,-25) -- (-23,20);
\draw[color = black!30] (-45,-45) -- (-45,0);
\draw[color = black!30] (90,-45) -- (90,0);
\draw[color = black!30] (-30,-35) -- (-30,0);
\draw[color = black!30] (30,-35) -- (30,0);


\draw[dashed, color = black] (-45,-5) -- (0,14.8);
\draw[dashed, color = black] (0,14.8) -- (90,10);

\draw[dashed, color = black] (-45,-5) -- (0,4);
\draw[dashed, color = black] (0,4) -- (90,10);

\draw[dashed, color = black] (-45,-5) -- (0,-6.8);
\draw[dashed, color = black] (0,-6.8) -- (90,10);

\draw[dashed, color = black] (-45,-5) -- (0,-17.6);
\draw[dashed, color = black] (0,-17.6) -- (90,10);

\LensBig{0}{0}{40}{black};
\filldraw (30,0) circle (20pt);
\filldraw (-30,0) circle (20pt);

\filldraw (-20,18) circle (20pt);
\filldraw (-20,12) circle (20pt);
\filldraw (-20,6) circle (20pt);
\filldraw (-20,0) circle (20pt);
\filldraw (-20,-6) circle (20pt);
\filldraw (-20,-12) circle (20pt);
\filldraw (-20,-18) circle (20pt);

\draw[line width = 1pt, color = black] (-23,-20) -- (-23,20);

\draw[line width = 1pt, color = drawRed] (90,0) -- (90,10);
\draw[line width = 1pt, color = drawRed] (-45,0) -- (-45,-5);

\draw[color = black] (-50,0) -- (95,0);

\draw[color = black,->] (-23.1,-25) -- (-23,-25);
\draw[color = black] (-19,-25) -- (-23.1,-25);
\node[below, black, opacity = 1] at(-21.5,-25) {\scriptsize $B$};
\draw[color = black, <->] (0,-25) -- (-20,-25);
\node[below, black, opacity = 1] at(-10,-25) {\scriptsize $b_{L0}$};
\draw[color = black, <->] (0,-35) -- (-30,-35);
\node[below, black, opacity = 1] at(-15,-35) {\scriptsize $f_L$};
\draw[color = black, <->] (0,-35) -- (30,-35);
\node[below, black, opacity = 1] at(15,-35) {\scriptsize $f_L$};
\draw[color = black, <->] (0,-45) -- (-45,-45);
\node[below, black, opacity = 1] at(-22.5,-45) {\scriptsize $b_L$};
\draw[color = black, <->] (0,-45) -- (90,-45);
\node[below, black, opacity = 1] at(45,-45) {\scriptsize $z_C$};

\node[above right, black, opacity = 1] at(-1,19) {\tiny main lens};
\node[above left, black, opacity = 1] at(-22,19) {\tiny sensor};
\node[above right, black, opacity = 1] at(-21,19) {\tiny MLA};

\node[above, black, opacity = 1] at(90,10) {\tiny object};
\node [above, black, label={[align=center,font=\tiny\linespread{0.5}] virtual\\image}] at(-45,-5){};
\end{tikzpicture}\label{fig:cross_view_plenoptic_camera}}\hfill
\subfigure[]{

\newcommand*\LensBig[4]{%
  \draw[color=#4] (#1,#2) ++(0,#3/2) arc(180-53.13010235:180+53.13010235:#3*0.625) arc(-53.13010235:53.13010235:#3*0.625) -- cycle ++(0,-#3)
	}

\begin{tikzpicture}[
		scale=0.045,
    >=stealth,
    inner sep=0pt, outer sep=2pt,%
]

\useasboundingbox (-70,-70) rectangle (100,38);

\LensBig{0}{0}{40}{black!30};
\filldraw (30,0) circle (20pt);
\filldraw (-30,0) circle (20pt);

\filldraw[color = black!30] (-20,18) circle (20pt);
\filldraw[color = black!30] (-20,12) circle (20pt);
\filldraw[color = black!30] (-20,6) circle (20pt);
\filldraw[color = black!30] (-20,0) circle (20pt);
\filldraw[color = black!30] (-20,-6) circle (20pt);
\filldraw[color = black!30] (-20,-12) circle (20pt);
\filldraw[color = black!30] (-20,-18) circle (20pt);

\draw[line width = 1pt, color = black!30] (-23,-20) -- (-23,20);

\draw[color = black!30] (-60,38) -- (-60,-57);
\draw[color = black!30] (0,-47) -- (0,20);
\draw[color = black!30] (90,-57) -- (90,0);


%
%
%

\draw[dashed, color = black] (-60,18) -- (90,10);
\draw[dashed, color = black] (-60,-0) -- (90,10);
\draw[dashed, color = black] (-60,-18) -- (90,10);
\draw[dashed, color = black] (-60,-36) -- (90,10);

\draw[dotted, color = black] (-60,36) -- (0,0);
\draw[dotted, color = black] (-60,18) -- (0,0);
\draw[dotted, color = black] (-60,-0) -- (0,0);
\draw[dotted, color = black] (-60,-18) -- (0,0);
\draw[dotted, color = black] (-60,-36) -- (0,0);

\filldraw (-60,36) circle (20pt);

\filldraw (-60,18) circle (20pt);

\filldraw (-60,0) circle (20pt);

\filldraw (-60,-18) circle (20pt);

\filldraw (-60,-36) circle (20pt);


\draw[line width = 1pt, color = black] (-57,-38) -- (-57,38);

\draw[line width = 1pt, color = drawRed] (90,0) -- (90,10);

\draw[color = black] (-63,0) -- (95,0);

\draw[color = black, <->] (0,-47) -- (90,-47);
\node[below, black, opacity = 1] at(45,-47) {\scriptsize $z_C$};
\draw[color = black, <->] (0,-47) -- (-60,-47);
\node[below, black, opacity = 1] at(-30,-47) {\scriptsize $\left|z_{C0}\right|$};
\draw[color = black, <->] (-60,-57) -- (90,-57);
\node[below, black, opacity = 1] at(15,-57) {\scriptsize $z_C'$};

\node[above, black!30, opacity = 1] at(0,19) {\tiny main lens};

\node[above, black, opacity = 1] at(90,10) {\tiny object};
\node[above, black, opacity = 1,rotate=90] at(-62,0) {\tiny virtual camera array};
\end{tikzpicture}\label{fig:cross_view_plenoptic_camera_multi_view}}
\caption{Projection model of a focused plenoptic camera. (a) Original plenoptic camera model. Main lens is represented by a thin lens, while the micro lenses in the \acs{mla} are pinholes which project the virtual image on the sensor. (b) plenoptic camera represented as an array of virtual cameras which observe directly the object space. The model in (b) represents a projection model equivalent to (a).}
\label{fig:plenoptic_camera_projection_model}
\end{figure}

The projection model which is used for the plenoptic camera in this dataset, is the one proposed in \cite{Zeller2017} and visualized in Figure~\ref{fig:plenoptic_camera_projection_model}.
In this model, the main lens of the plenoptic camera is a thin lens, while the micro lenses in the \ac{mla} are pinholes.
In \cite{Zeller2017} it was shown, that this model forms, in fact, the equivalent to a virtual camera array, as it is shown in Figure~\ref{fig:cross_view_plenoptic_camera_multi_view}, where each micro image represents the image of a small pinhole camera with a very narrow \acl{fov}.

Using this model, one obtains the coordinates of a point $\boldsymbol x_p$ in a virtual camera from the camera coordinates $\boldsymbol x_C$ of a 3D object point as follows:
\begin{align}
z_C' \begin{bmatrix}
x_p\\y_p\\1
\end{bmatrix} = z_C' \boldsymbol x_p = 
\boldsymbol{x}_C - \boldsymbol{p}_{ML} = 
\begin{bmatrix}
x_C\\y_C\\z_C
\end{bmatrix} -
\begin{bmatrix}
p_{MLx}\\p_{MLx}\\-z_{C0}
\end{bmatrix}.
\end{align}
Here, $\boldsymbol p_{ML}$ defines the center of the virtual camera, or projected micro lens.
The center $\boldsymbol p_{ML}$ is calculated from the center of the real micro lens $\boldsymbol c_{ML}$ as follows:
\begin{align}
\boldsymbol{p}_{ML} &= 
\begin{bmatrix}
p_{MLx}\\p_{MLy}\\-z_{C0}
\end{bmatrix}
= -\boldsymbol{c}_{ML} \frac{z_{C0}}{b_{L0}} = 
-\begin{bmatrix}
c_{MLx}\\c_{MLy}\\b_{L0}
\end{bmatrix}
\frac{z_{C0}}{b_{L0}} \nonumber\\
&= -\boldsymbol{c}_{ML} \frac{f_L}{f_L-b_{L0}} = \boldsymbol{c}_{ML} \frac{f_L}{b_{L0}-f_L}.
\label{eq:projected_micro_lens}
\end{align}
The parameter $z_{C0}$ defines the distance from the real main lens of the plenoptic camera to the virtual camera array (see Fig.~\ref{fig:cross_view_plenoptic_camera_multi_view}).
\begin{align}
z_{C0} &:= \frac{f_L \cdot b_{L0}}{f_L-b_{L0}}
\end{align}

Furthermore, a point $\boldsymbol x_{ML}$ in a real micro image can be calculated from the corresponding point $\boldsymbol x_{p}$ in the respective virtual camera as follows:
\begin{align}
\boldsymbol x_{ML} = \begin{bmatrix}
x_{ML}\\y_{ML}\\B
\end{bmatrix} = \boldsymbol x_p \cdot \frac{f_L \cdot B}{f_L-b_{L0}} - \boldsymbol c_{ML} \cdot \frac{B}{f_L-b_{L0}}.
\label{eq:projected_micro_lens_4_new}
\end{align}
The micro image point $\boldsymbol x_{ML}$, given in eq.~\eqref{eq:projected_micro_lens_4_new}, is an image point relative to its micro lens center $\boldsymbol c_{ML}$.
Hence, corresponding raw image coordinates $\boldsymbol x_R$, which are unique for each single point in the entire raw image recorded by the plenoptic camera, are defined:
\begin{align}
\boldsymbol x_R =
\begin{bmatrix}
x_R\\y_R
\end{bmatrix} =  
\begin{bmatrix}
x_{ML}\\y_{ML}
\end{bmatrix} +
\begin{bmatrix}
c_{MLx}\\c_{MLy}
\end{bmatrix}.
\end{align}

The common way to obtain the centers of the micro lenses in the \ac{mla} is to estimate them based on a recorded white image \cite{Dansereau2013}.
However, these centers, in fact, do not represent the micro lens centers $\boldsymbol c_{ML}$, but instead the corresponding micro image centers $\boldsymbol c_{I}$.
As it is shown in \cite{Zeller2018a}, the micro lens center $\boldsymbol c_{ML}$ and the micro image centers $\boldsymbol c_{I}$ have the following relationship:
\begin{align}
\boldsymbol c_{ML} =
\begin{bmatrix}
c_{MLx}\\c_{MLy}\\b_{L0}
\end{bmatrix} := \boldsymbol c_I \frac{b_{L0}}{b_{L0}+B} =
\begin{bmatrix}
c_{Ix}\\c_{Iy}\\b_{L0}+B
\end{bmatrix} \frac{b_{L0}}{b_{L0}+B}.
\end{align}

To correct for lens distortions, a distortion model is applied to the raw image coordinates $\boldsymbol x_R$.
Hence, the following connection between the distorted coordinates $\boldsymbol x_{Rd}$, which, in fact, are the coordinates of the image recorded by the camera, and the undistorted coordinates $\boldsymbol x_R$ is defined:
\begin{align}
\boldsymbol x_{Rd} = 
\begin{bmatrix}
x_{Rd}\\y_{Rd}
\end{bmatrix} = 
\begin{bmatrix}
x_{R}\\y_{R}
\end{bmatrix} + 
\begin{bmatrix}
\Delta x_\text{dist}\\\Delta y_\text{dist}
\end{bmatrix}.
\end{align}
Here, the distortion terms, $\Delta x_\text{dist}$ and $\Delta y_\text{dist}$, consist of a radial symmetric as well as a tangential distortion component and are defined as follows:
\begin{align}
\Delta x_\text{dist} &= x_R (A_0 r^2 + A_1 r^4) +  B_0 \cdot \left(r^2 + 2 x_R^2\right) + 2 B_1 x_R y_R,\\
\Delta y_\text{dist} &= y_R (A_0 r^2 + A_1 r^4) + B_1 \cdot \left(r^2 + 2 y_R^2\right) + 2 B_0 x_R y_R,\\
r &= \sqrt{x_R^2 + y_R^2}.
\end{align}
For the plenoptic camera, we found two radial symmetric parameters ($A_0$, and $A_1$) to be sufficient to model the distortion.

The micro image centers are detected also on distorted raw image coordinates $\boldsymbol c_{Id}$ and therefore have to be corrected by the same distortion model.
Hence, the corrected micro image centers $\boldsymbol c_{I}$ will not be arranged on a regular hexagonal grid anymore, but will slightly deviate from this grid.

So far, the coordinates $\boldsymbol x_{Rd}$ were defined in metric dimension and relative to the optical axis. 
Hence, they still have to be transformed into pixel coordinates $\boldsymbol x_{Rd}'$ as follows:
\begin{align}
\boldsymbol x_{Rd}' = 
\begin{bmatrix}
x_{Rd}'\\y_{Rd}'
\end{bmatrix} = 
\begin{bmatrix}
x_{Rd}\\y_{Rd}
\end{bmatrix} \cdot s^{-1} +
\begin{bmatrix}
c_x\\c_y
\end{bmatrix}.
\end{align}
Here, $s$ defines the size of a pixel and $\boldsymbol c = [c_x,c_y]^T$ is the so-called principal point.

As already mentioned, the micro image centers $\boldsymbol c_{Id}$ can be estimated from a recorded white image.
Furthermore, the pixel size $s$ can be taken directly from the sensor specifications.
All other parameter, have to be estimated in a geometric calibration.
These parameters are:
\begin{itemize} 
\item[$\bullet$] main lens focal length: $f_L$
\item[$\bullet$] distance between main lens and \ac{mla}: $b_{L0}$
\item[$\bullet$] distance between \ac{mla} and sensor: $B$
\item[$\bullet$] principal point (in pixels): $\boldsymbol c = [c_x,c_y]^T$
\item[$\bullet$] four distortion parameters: $A_0$, $A_1$, $B_0$, and $B_1$
\end{itemize}

\subsubsection{Stereo Camera Model}
For the two monocular cameras in the stereo setup we define the pinhole camera model as given in eq.~\eqref{eq:stereo_camera_model}.
\begin{align}
\lambda \begin{bmatrix}
x_I\\y_I\\1
\end{bmatrix} = 
\begin{bmatrix}
f_x&0&c_x\\
0&f_y&c_y\\
0&0&1
\end{bmatrix} \cdot
\begin{bmatrix}
x_C\\y_C\\z_C
\end{bmatrix}
\label{eq:stereo_camera_model}
\end{align}
In contrast to the plenoptic camera model, we define different focal lengths ($f_x$, $f_y$) in $x$- and $y$-direction.
Thereby, we are able to consider rectangular, instead of squared, sensor pixels.

Lens distortion is applied on normalized image coordinates as given in eq.~\eqref{eq:lens_distortion_stereo_camera}.
\begin{align}
\boldsymbol x_{Id} = 
\begin{bmatrix}
x_{Id}\\
y_{Id}
\end{bmatrix} =
\begin{bmatrix}
f_x(x + \Delta x_\text{dist}) +c_x\\
f_y(y + \Delta y_\text{dist}) +c_y
\end{bmatrix} \quad \text{with} \quad \text{$x = \frac{x_C}{z_C}$ and $y = \frac{y_C}{z_C}$}
\label{eq:lens_distortion_stereo_camera}
\end{align}
Due to the much larger \ac{fov}, rather than two, we have to consider three parameters for radial symmetric distortion.
Furthermore, the effect of tangential distortion is negligible.
\begin{align}
\Delta x_\text{dist} &= x (A_0 r^2 + A_1 r^4 + A_2 r^6)\label{eq:lens_distortion_stereo_camera_x}\\
\Delta y_\text{dist} &= y (A_0 r^2 + A_1 r^4 + A_2 r^6)\label{eq:lens_distortion_stereo_camera_y}
\end{align}
The variable $r = \sqrt{x^2+y^2}$ defines the distance to the principal point on the sensor in normalized image coordinates.
In addition to the intrinsic parameters, the orientation of the slave (right) camera with respect to the master (left) camera is defined by a rigid body transformation $\boldsymbol G(\boldsymbol \xi_{MS}) \in \mathrm{SE(3)}$, which is represented by the respective tangent space element $\boldsymbol \xi_{MS} \in \mathfrak{se}(3)$.

\subsubsection{Calibration Approach}\label{sec:calibration_approach}
\begin{figure}[t]
\centering
\subfigure[3D calibration target]{\includegraphics[width=0.48\linewidth]{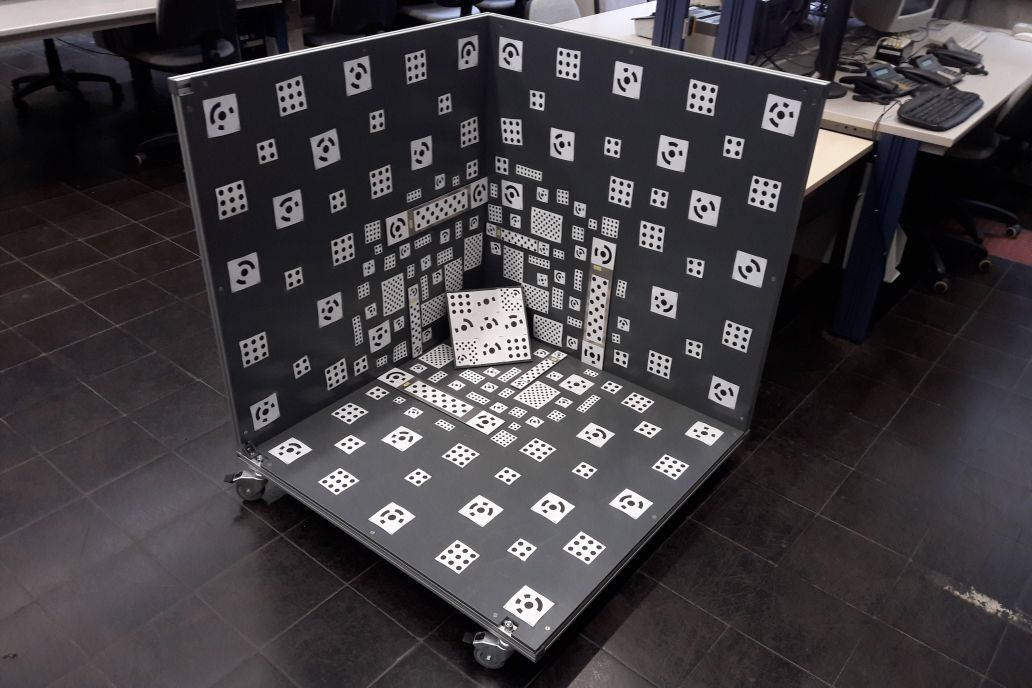}\label{fig:calibration_setup_3d_calibration_target}}\hfill
\subfigure[white balance filter]{\includegraphics[width=0.48\linewidth]{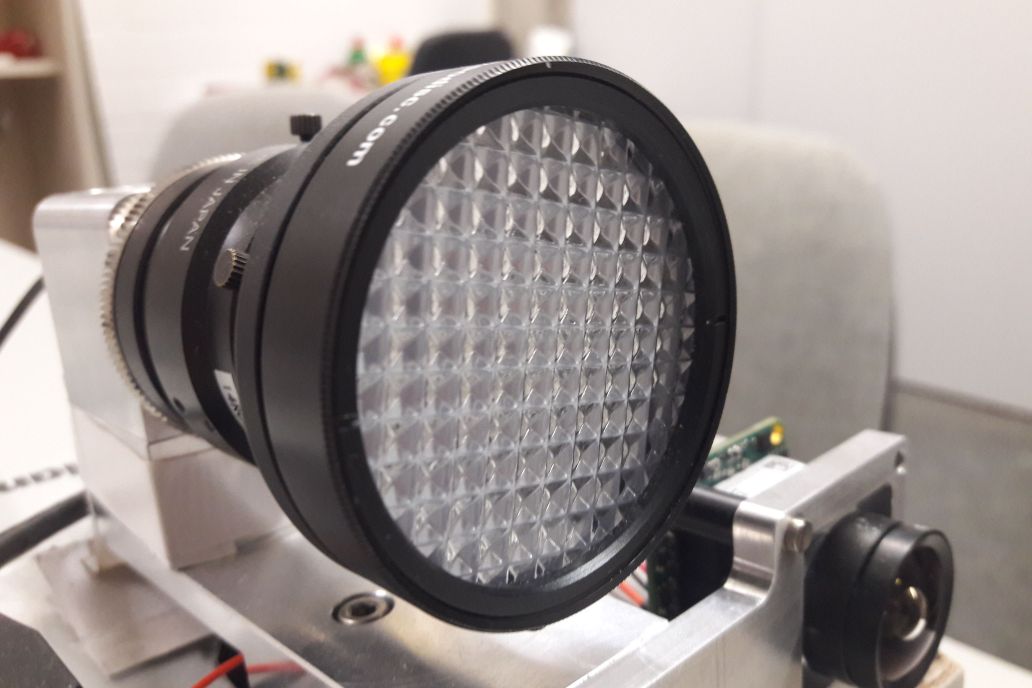}\label{fig:white_balance_filter}}
\caption{Camera calibration setup. (a) 3D calibration target used for geometric calibration of the plenoptic camera and the stereo camera system. (b) White balance filter used to record white images for the plenoptic camera and the stereo camera system.
The white images are used for vignetting correction.}

\end{figure}

For both systems, the plenoptic camera and the stereo camera, the model parameters are estimated from a set of images in a full bundle adjustment.
For this purpose we use the \acs{3d} calibration target shown in Figure~\ref{fig:calibration_setup_3d_calibration_target}.
In the bundle adjustment, all parameters of the camera models, the extrinsic orientations of the single images as well as the 3D coordinates of the calibration markers are estimated.

The micro images of the plenoptic camera generally do not cover a complete marker point.
Therefore, the marker points cannot be detected reliably in the micro images.
Hence, we calculate from each raw image, recorded by the plenoptic camera, the corresponding totally focused image.
Afterwards, the marker points are detected in the totally focused image and are projected back to the micro images in the respective raw image.
This procedure was described already in \cite{Zeller2017}.

\subsection{Vignetting Correction}\label{sec:vignetting_correction}

Especially for direct \ac{vo} approaches, vignetting has a negative effect on the performance.
The model parameters are estimated based on photometric measurements and therefore the measured intensity of a point must be independent of its location on the sensor.
Indirect methods are more robust to vignetting, as extracted feature points generally rely on corners in the images, which are invariant to absolute intensity changes.

While in a monocular camera vignetting generally results in a continuously increasing attenuation of pixel intensities from the image center towards the boundaries, in a plenoptic camera further vignetting effects are present for each single micro lens.

Mathematically we can describe the vignetting as follows:
\begin{align}
I(\boldsymbol x) = \tau \cdot \left(V(\boldsymbol x) \cdot B(\boldsymbol x)+ \epsilon_I\right).
\end{align}
The function $I(\boldsymbol x)$ is the observed intensity value measured by the sensor, while $B(\boldsymbol x)$ is the irradiance image which represents the scene in a photometrically correct way.
The vignetting $V(\boldsymbol x)$ defines a pixel-wise attenuation, with $V(\boldsymbol x) \in [0,1]$.
We use the notations $\tau$ for the exposure time and $\epsilon_I$ for the sensor noise.
In this simplified model, we considered the image sensor to have a linear transfer characteristic which is, in fact, not the case for a real sensor.

\begin{figure}[t]
\centering
\subfigure[]{\begin{tikzpicture}

\begin{axis}[%
width=.4\linewidth,
scale only axis,
point meta min=0.5,
point meta max=1,
axis on top,
xmin=0.5,
xmax=2048.5,
xtick={\empty},
y dir=reverse,
ymin=0.5,
ymax=2048.5,
ytick={\empty},
axis background/.style={fill=white},
colormap/hot2,
colorbar
]
\addplot [forget plot] graphics [xmin=0.5,xmax=2048.5,ymin=0.5,ymax=2048.5] {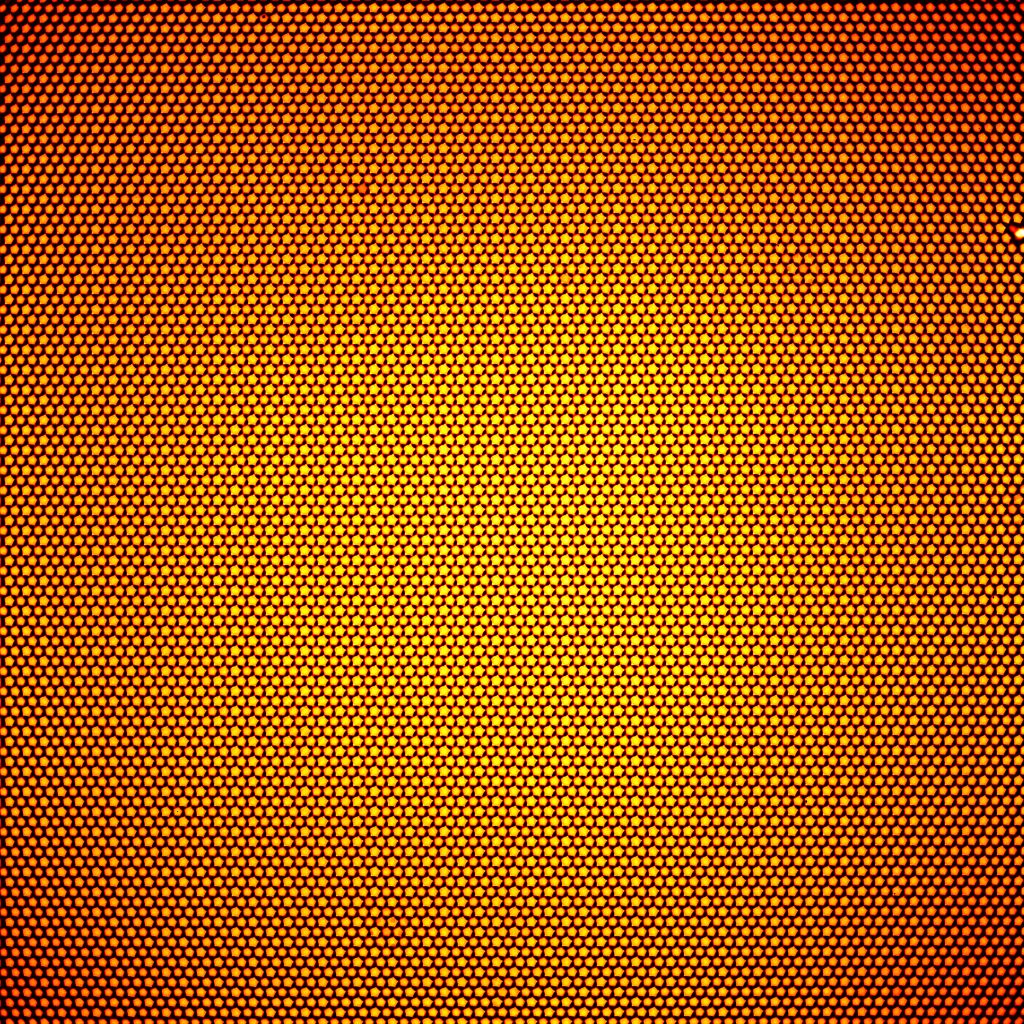};
\end{axis}
\end{tikzpicture}%
}\hfill
\subfigure[]{\begin{tikzpicture}
\begin{axis}[%
width=.4\linewidth,
scale only axis,
xmin=0.0,
xmax=2047,
xtick={\empty},
ymin=0.5,
ymax=1,
ylabel={attenuation factor},
ylabel near ticks,
]
\addplot[color=gColor1, mark=none] table [x expr=\thisrowno{0},y expr=\thisrowno{1},col sep=comma]{figures/vignetting/virgnetting_plenoptic.csv};
\end{axis}
\end{tikzpicture}%
}
\caption[Estimated attenuation image for the plenoptic camera]{Estimated attenuation image for the plenoptic camera. (a) Complete attenuation image. (b) Horizontal cross section through the attenuation map. Vignetting is visible in each micro image as well as across the complete image resulting from the main lens.}
\label{fig:plenoptic_white_image}
\end{figure}
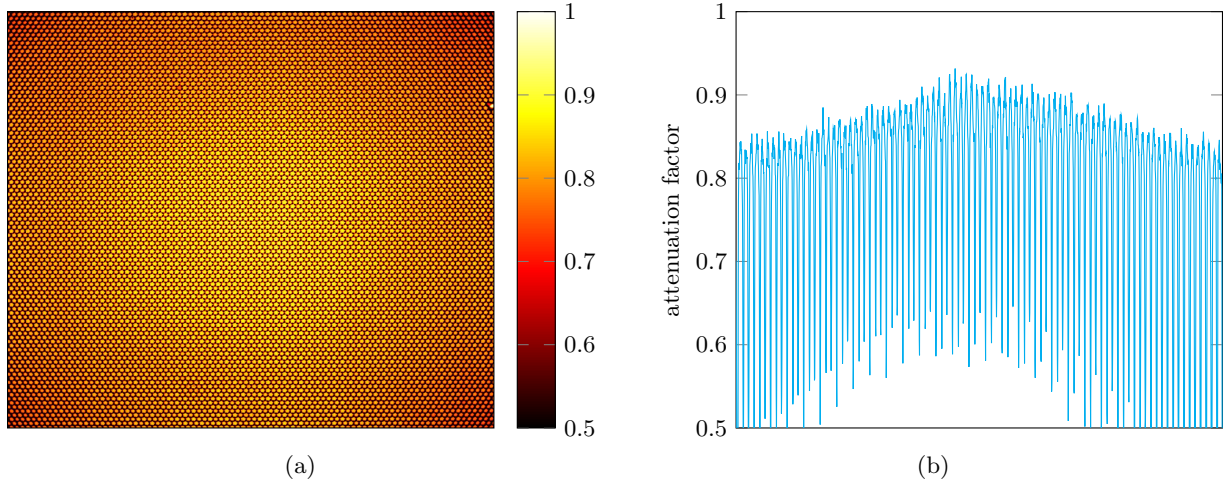

While the nonparametric vignetting compensation based on white images, recorded with a white balance filter, is commonly applied to plenoptic cameras, we correct the vignetting of the two cameras in the stereo system in the same way.
For each camera we recorded a set of 10 white images and calculated an average attenuation image from this set.
Figure~\ref{fig:white_balance_filter} shows the used white balance filter.
For the two cameras of the stereo system, we additionally filtered the attenuation images using a Gaussian kernel.
This cannot be done for the attenuation image of the plenoptic camera, as in this case, the \ac{mla} produces quite high frequent components in the attenuation image which must be preserved (see Figure~\ref{fig:plenoptic_white_image}).

In \cite{Engel2016a} a different method is described, where the attenuation map is calculated based on a sequence of images capturing a white wall.
However, the method \cite{Engel2016a} is quite time consuming and, in our experience, error prone\footnote{Reflections and shadows on the white wall negatively affect the results.}.
For the method \cite{Engel2016a}, one has to capture a sequence of hundreds of images and then run the estimation for up to one hour.
The attenuation map based on the white balance filter, by contrast, is obtained in just a few seconds.
Furthermore, we want to apply comparable calibrations to both systems; the plenoptic camera as well as the stereo cameras.

Figure~\ref{fig:plenoptic_white_image} shows the vignetting for the plenoptic camera. 
As one can see, vignetting is visible in each individual micro image.
Furthermore, outer image regions are attenuated stronger than the image center.
This is due to the influence of the main lens.
In addition, there are some small irregularities visible in the map resulting from defect micro lenses and dirt on the \ac{mla}.

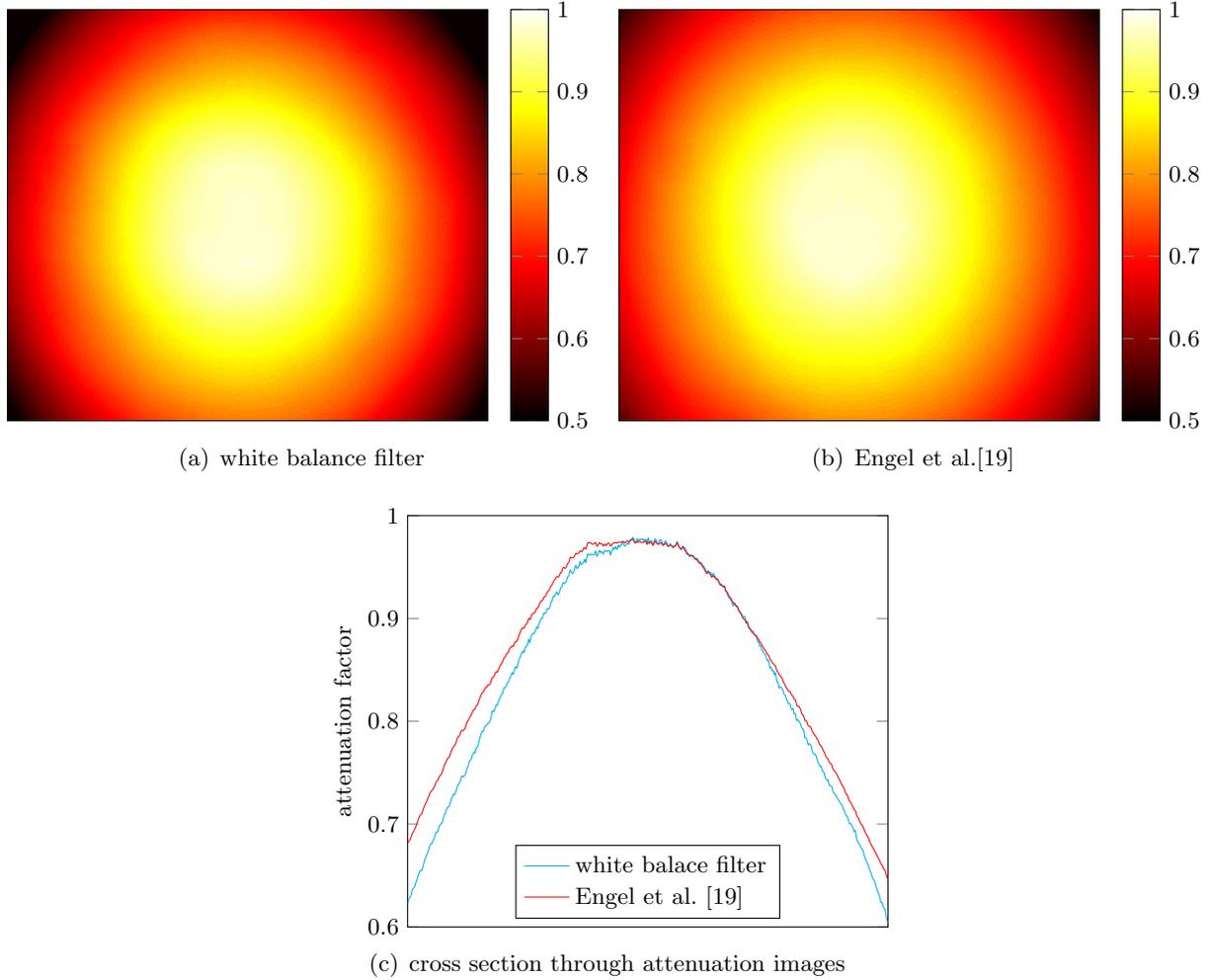
\begin{figure}[t]
	\centering
	\subfigure[white balance filter]{\begin{tikzpicture}

\begin{axis}[%
width=.4\linewidth,
scale only axis,
point meta min=0.5,
point meta max=1,
axis on top,
xmin=0.5,
xmax=1280.5,
xtick={\empty},
y dir=reverse,
ymin=0.5,
ymax=1024.5,
ytick={\empty},
axis background/.style={fill=white},
colormap/hot2,
colorbar
]
\addplot [forget plot] graphics [xmin=0.5,xmax=1280.5,ymin=0.5,ymax=1024.5] {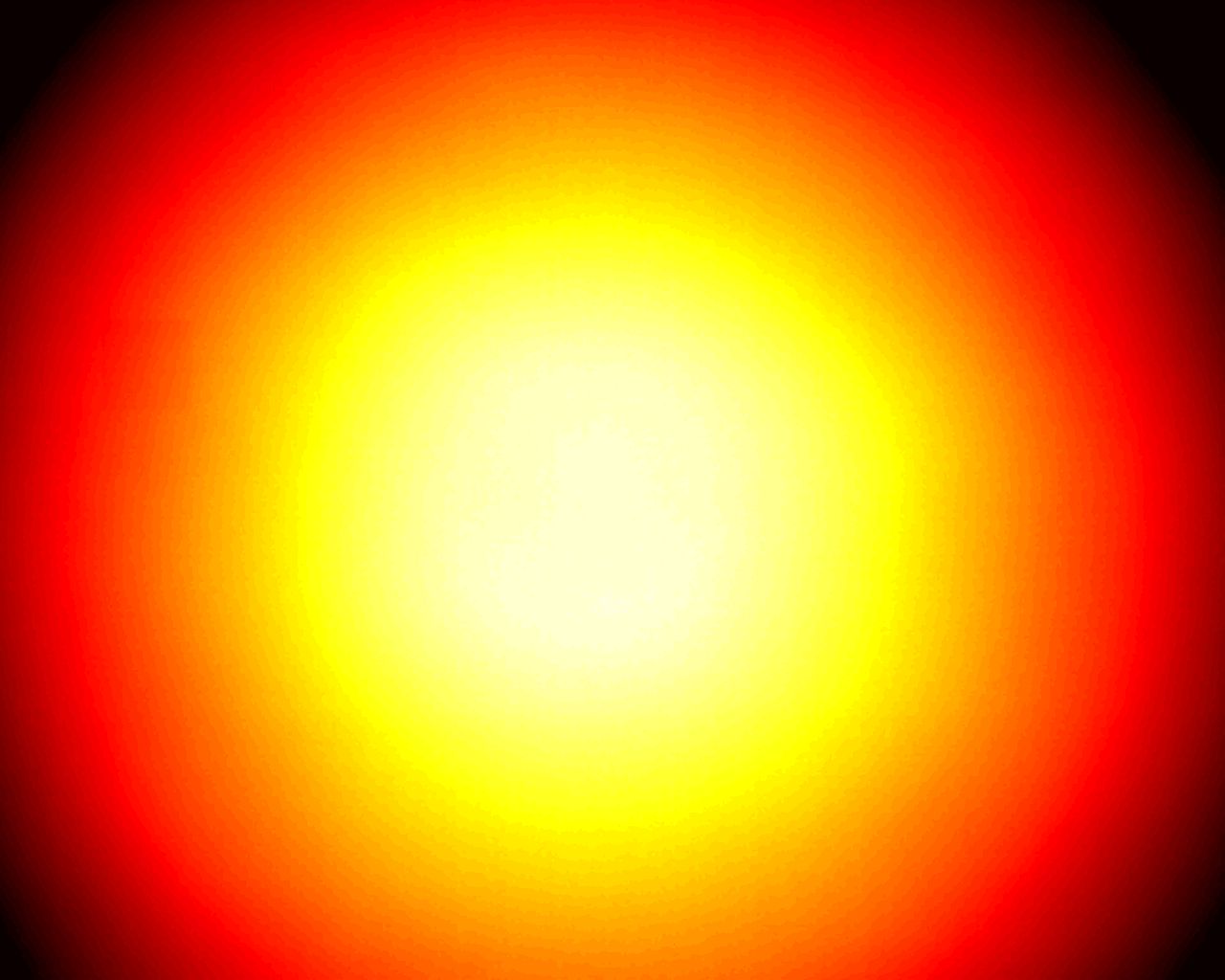};
\end{axis}
\end{tikzpicture}%
\label{fig:stereo_white_image_01}}\hfill
	\subfigure[Engel et al.\cite{Engel2016a}]{\begin{tikzpicture}

\begin{axis}[%
width=.4\linewidth,
scale only axis,
point meta min=0.5,
point meta max=1,
axis on top,
xmin=0.5,
xmax=1280.5,
xtick={\empty},
y dir=reverse,
ymin=0.5,
ymax=1024.5,
ytick={\empty},
axis background/.style={fill=white},
colormap/hot2,
colorbar
]
\addplot [forget plot] graphics [xmin=0.5,xmax=1280.5,ymin=0.5,ymax=1024.5] {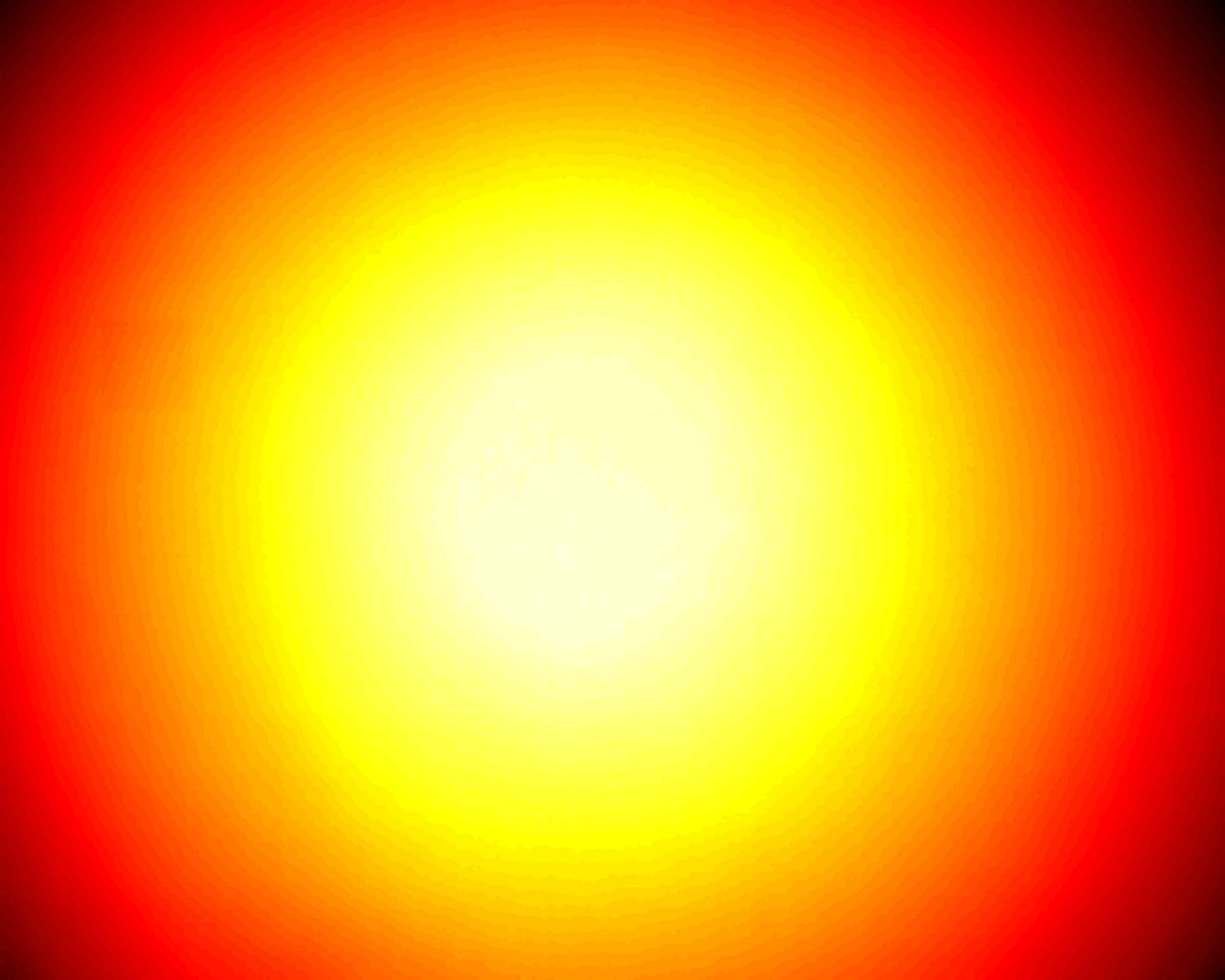};
\end{axis}
\end{tikzpicture}%
\label{fig:stereo_white_image_02}}\\
	\subfigure[cross section through attenuation images]{\begin{tikzpicture}
\begin{axis}[%
width=.4\linewidth,
scale only axis,
xmin=0.0,
xmax=1279,
xtick={\empty},
ymin=0.6,
ymax=1,
ylabel={attenuation factor},
ylabel near ticks,
legend style={at={(0.5,0.02)}, anchor=south, legend cell align=left}
]
\addplot[color=gColor1, mark=none] table [x expr=\thisrowno{0},y expr=\thisrowno{1},col sep=comma]{figures/vignetting/virgnetting_cam0.csv};
\addlegendentry{white balace filter};
\addplot[color=gColor2, mark=none] table [x expr=\thisrowno{0},y expr=\thisrowno{2},col sep=comma]{figures/vignetting/virgnetting_cam0.csv};
\addlegendentry{Engel et al. \cite{Engel2016a}};  
\end{axis}
\end{tikzpicture}%
\label{fig:stereo_white_image_03}}
	\caption[Estimated attenuation images for the left camera of the stereo camera system]{Estimated attenuation images for the left camera of the stereo camera system. (a) Attenuation image recorded with the white balance filter. (b) Attenuation image calculated based on the method described in \cite{Engel2016a}. (c) Horizontal cross section through the resulting attenuation images.}
	\label{fig:stereo_white_image}
\end{figure}

Figure~\ref{fig:stereo_white_image} shows the attenuation maps estimated for the left camera of the stereo system.
Figure~\ref{fig:stereo_white_image_01} shows the result using the white balance filter, while Figure~\ref{fig:stereo_white_image_02} shows the one obtained from the method of Engel et al. \cite{Engel2016a}.
From Figure~\ref{fig:stereo_white_image_03} one can clearly see that for the white balance filter the attenuation is sightly stronger at the sensor boundaries than for the method in \cite{Engel2016a}.
However, the deviation is quite small and furthermore, it is difficult to evaluate which map describes the vignetting of the camera in a more accurate way.

\subsection{Ground Truth and Evaluation Metric}\label{sec:ground_truth_and_evaluation_metric}
\begin{figure}[t]
\centering
\includegraphics[width=\linewidth]{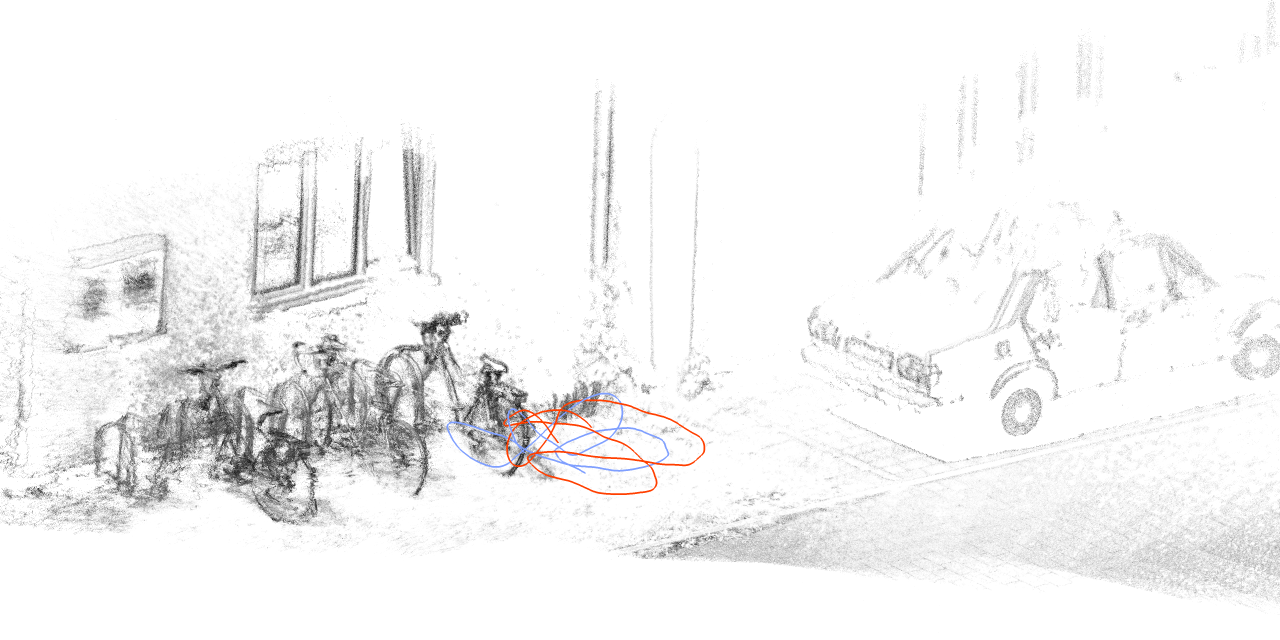}
\caption{Semi-dense point cloud generated by \cite{Zeller2018a} overlaid with the loop closure trajectory of the start and end segment.
The loop closure trajectory is used as ground truth and is obtained from \acs*{sfm} based on the stereo images.
Start segment of the trajectory is marked in red and the end segment in blue.}
\label{fig:point_cloud_gt_data}
\end{figure}

It is almost impossible to obtain ground truth trajectories for long and large-scale sequences recorded by hand-held cameras.
We decided to obtain ground truth data for our dataset in a similar way as suggested by Engel et al. \cite{Engel2016a}, where the accuracy of a \ac{vo} algorithm is evaluated based on a single, larger loop closure.
Each trajectory in the dataset starts with a winding sequence while capturing a nearby object.
This starting sequence is followed by the actual trajectory which finally leads back to the starting point in a large loop, followed by a short, winding, finishing sequence.

Using the winding sequence at the beginning and at the end, we are able to register both segments to each other using a standard \ac{sfm} approach.
These registered segments then can be used as ground truth information.
In contrast to monocular datasets, we also want ground truth data for the absolute scale of the trajectory.
The absolute scale of the trajectory is obtained from stereo images.
Since the stereo cameras have a much larger stereo baseline than the micro images in the plenoptic camera, the observed scale is accurate enough to serve as reference for the plenoptic sequences.

While one can basically use any \ac{sfm} algorithm to register the start and end segment to each other, we use a modified version of ORB-SLAM2 (stereo) \cite{Mur-Artal2017}.
Instead of selecting keyframes, we build up a frame-wise pose graph which is optimized in a global bundle adjustment.

Figure~\ref{fig:point_cloud_gt_data} shows, by way of example, the registered segments for the beginning and the end of the sequence in blue and red respectively, overlaid with the point cloud calculated by \cite{Zeller2018a}.

Using the registered ground truth data, based on each recorded trajectory two similarity transformations  $\boldsymbol T_s^\text{gt}$ and $\boldsymbol T_e^\text{gt}$ ($\in \text{Sim(3)}$) with respect to the start and the end segment of the sequence can be calculated, as given in eqs.~\eqref{eq:qt_registration_start} and~\eqref{eq:qt_registration_end}.
\begin{align}
\boldsymbol T_s^\text{gt}&:=\argmin_{\boldsymbol T \in \text{Sim(3)}} \sum_{i\in S}(\boldsymbol T \boldsymbol p_i - \boldsymbol p_i^\text{gt})^2\label{eq:qt_registration_start}\\
\boldsymbol T_e^\text{gt}&:=\argmin_{\boldsymbol T \in \text{Sim(3)}} \sum_{i\in E}(\boldsymbol T \boldsymbol p_i - \boldsymbol p_i^\text{gt})^2\label{eq:qt_registration_end}
\end{align}
The vectors $\boldsymbol p_i \in \mathbb{R}^3$ are the estimated points of the trajectory while  $\boldsymbol p_i^\text{gt} \in \mathbb{R}^3$ are the respective points of the ground truth.
$S$ and $E$ define the sets of indices of the start end and segment respectively.
One may notice that in eqs.~\eqref{eq:qt_registration_start} and~\eqref{eq:qt_registration_end} actually the homogeneous representations of 3D points $\boldsymbol p_i$ and $\boldsymbol p_i^\text{gt}$ have to be used.

Using the similarity transformations $\boldsymbol T_s^\text{gt}$ and $\boldsymbol T_e^\text{gt}$ we define evaluation metrics similar to \cite{Engel2016a}.
From the two transformations the accumulated drift $\boldsymbol T_\text{drift} \in \mathrm{Sim(3)}$ from the start to the end of the trajectory can be calculated as follows:
\begin{align}
\boldsymbol T_\text{drift} :=
\begin{bmatrix}
e_s \boldsymbol R & \boldsymbol t\\
\boldsymbol 0 & 1
\end{bmatrix} = \boldsymbol T_e^\text{gt}(\boldsymbol T_s^\text{gt})^{-1} = 
\begin{bmatrix}
s_e \boldsymbol R_e & \boldsymbol t_e\\
\boldsymbol 0 & 1
\end{bmatrix}
\begin{bmatrix}
s_s \boldsymbol R_s & \boldsymbol t_s\\
\boldsymbol 0 & 1
\end{bmatrix}^{-1}.
\end{align}
From $\boldsymbol T_\text{drift}$ we can directly extract the scale drift $e_s$, the rotational drift $e_r$, and the translation drift $e_t := \Vert\boldsymbol t\Vert$.
The rotational drift $e_r$ is defined by the rotation angle around the Euler axis $\boldsymbol w \mapsto \widehat{\boldsymbol w}$ corresponding to the rotation matrix $\boldsymbol R \in \mathrm{SO}(3)$:
\begin{align}
e_r := \left\Vert \boldsymbol w\right\Vert \cdot \frac{\ang{180}}{\pi} \qquad \text{with $\boldsymbol w \mapsto \widehat{\boldsymbol w} = \log_{\mathrm{SO(3)}}(\boldsymbol R)$}.
\end{align}
The mapping $\boldsymbol w \mapsto \widehat{\boldsymbol w}$ defines the mapping of the vector $\boldsymbol w \in \mathbb{R}^3$ to the skew-symmetric matrix $\widehat{\boldsymbol w} \in \mathfrak{so}(3)$:
\begin{align}
\boldsymbol w =
\begin{bmatrix}
w_1\\w_2\\w_3
\end{bmatrix} \mapsto \widehat{\boldsymbol w} =
\begin{bmatrix}
0 & -w_3 & w_2\\w_3 & 0 & -w_1\\-w_2 & w_1 & 0
\end{bmatrix}.
\end{align}
For an easier interpretation of the scale drift $e_s' := \max\{e_s,e_s^{-1}\}$ is defined.

\begin{figure}[t]
\subfigure[sample images from the sequence of the left camera in the stereo camera system]{%
\includegraphics[width=.2\linewidth]{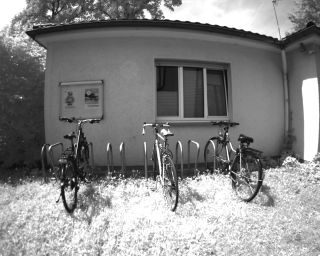}%
\includegraphics[width=.2\linewidth]{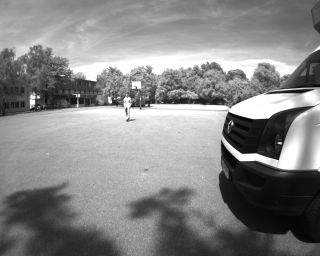}%
\includegraphics[width=.2\linewidth]{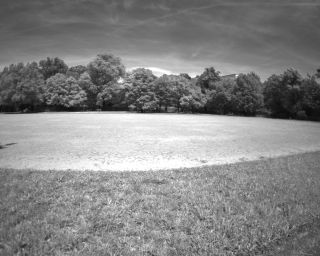}%
\includegraphics[width=.2\linewidth]{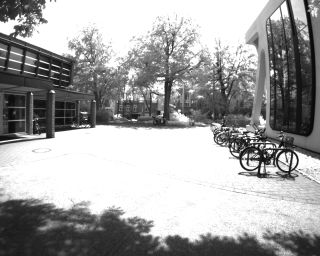}%
\includegraphics[width=.2\linewidth]{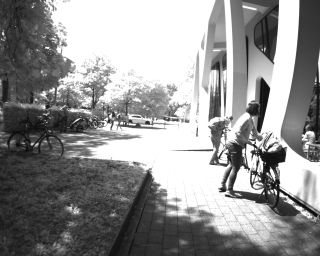}\label{fig:sample_images_sequence_stereo}
}

\subfigure[sample images from the sequence of the plenoptic camera]{%
\includegraphics[width=.2\linewidth]{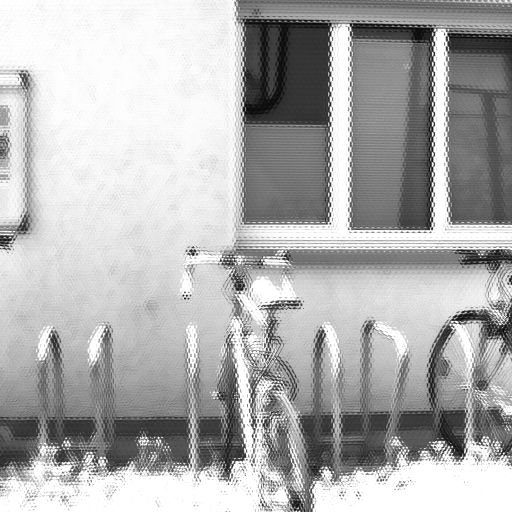}%
\includegraphics[width=.2\linewidth]{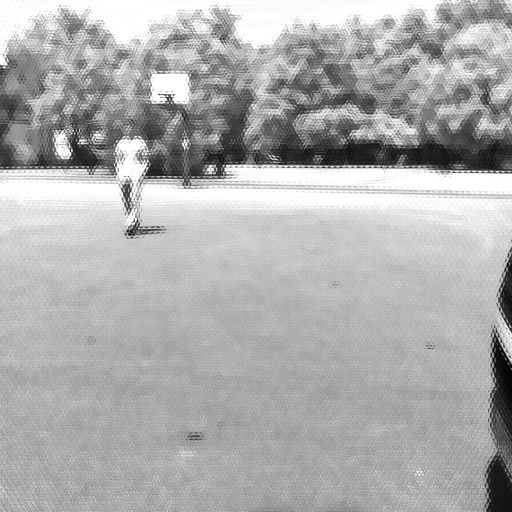}%
\includegraphics[width=.2\linewidth]{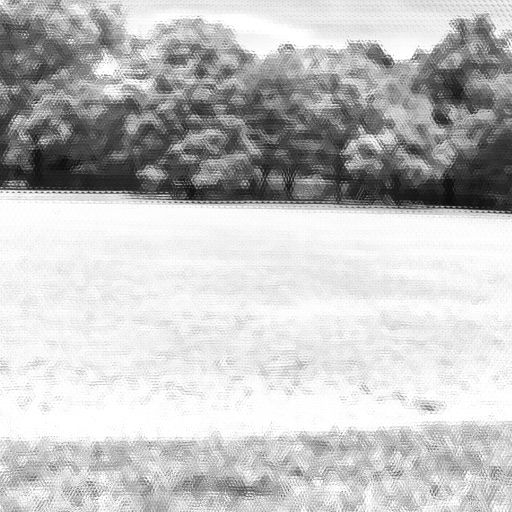}%
\includegraphics[width=.2\linewidth]{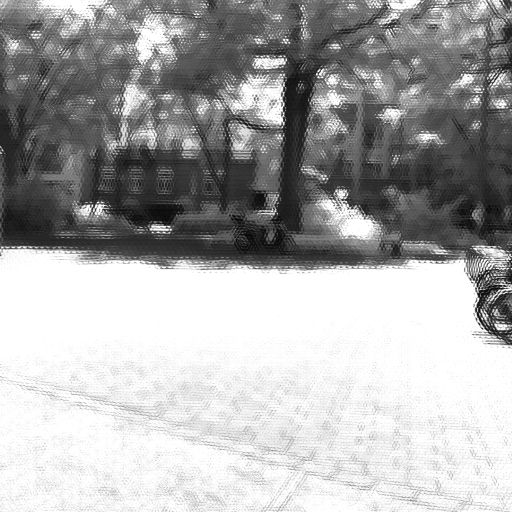}%
\includegraphics[width=.2\linewidth]{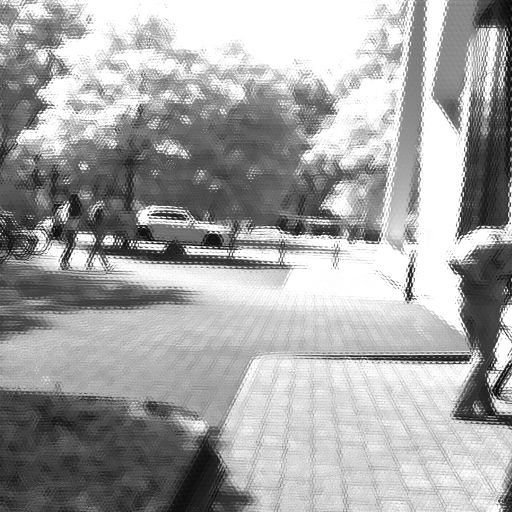}\label{fig:sample_images_sequence_plenoptic}
}
\caption[Sample images for one sequence of the synchronized stereo and plenoptic \acs*{vo} dataset]{Sample images for one sequence of the synchronized stereo and plenoptic \acs*{vo} dataset. (a) Sample image from the left camera of the stereo camera system.
(b) Sample images from the plenoptic camera. The images of both cameras correspond to exactly the same point in time.}
\label{fig:sample_images_sequence}
\end{figure}

The drift metrics $e_s'$, $e_r$, and $e_t$ define more or less independent quality measures.
Looking at just one of these values offers only a very limited insight into the overall quality of the estimated trajectory.
Furthermore, $e_t$, as it is defined here, is proportional to the absolute scale of the estimated trajectory and therefore is not meaningful at all without considering the absolute scale.
In \cite{Engel2016a} the alignment error $e_\text{align}$, as a more meaningful combined metric, is defined:
\begin{align}
e_\text{align} := \sqrt{\frac{1}{N} \sum_{i=1}^N{\left\Vert \boldsymbol T_s^\text{gt} \boldsymbol p_i - \boldsymbol T_e^\text{gt} \boldsymbol p_i \right\Vert_2^2}}.
\end{align}
The parameter $N$ is the number of points (frames) in the complete trajectory.
In comparison to $e_t$, $e_\text{align}$ is always scaled with respect to the ground truth and implicitly incorporates all drifts $e_s'$, $e_r$, $e_t$ in a single number.

All previously defined metrics consider only the relative drift from the beginning to the end of the trajectory, but not the error of the absolute scale.
We define the absolute scale difference $d_s$ of the front and end segment as another metric:
\begin{align}
d_s := \sqrt{\text{scale}(\boldsymbol T_e^\text{gt} \boldsymbol T_s^\text{gt})} = \sqrt{s_e \cdot s_s}.
\end{align}
Similar to the scale drift $e_s'$, we define $d_s' := \max\{d_s,d_s^{-1}\}$.

Of course, $d_s$ must be considered only for plenoptic and stereo algorithms and not for monocular approaches.
Furthermore, $d_s$ has significance only in combination with the scale drift $e'_s$:
\begin{align}
s_\text{max} &= d_s \cdot \sqrt{e'_s},\\
s_\text{min} &= \frac{d_s}{\sqrt{e'_s}}.
\end{align}

\begin{figure}[t]
\centering
\subfigure[top view real scene]{\begin{tikzpicture}\node[inner sep=0pt, anchor=north west,] (image)  at (0,0) {\includegraphics[width=0.48\linewidth]{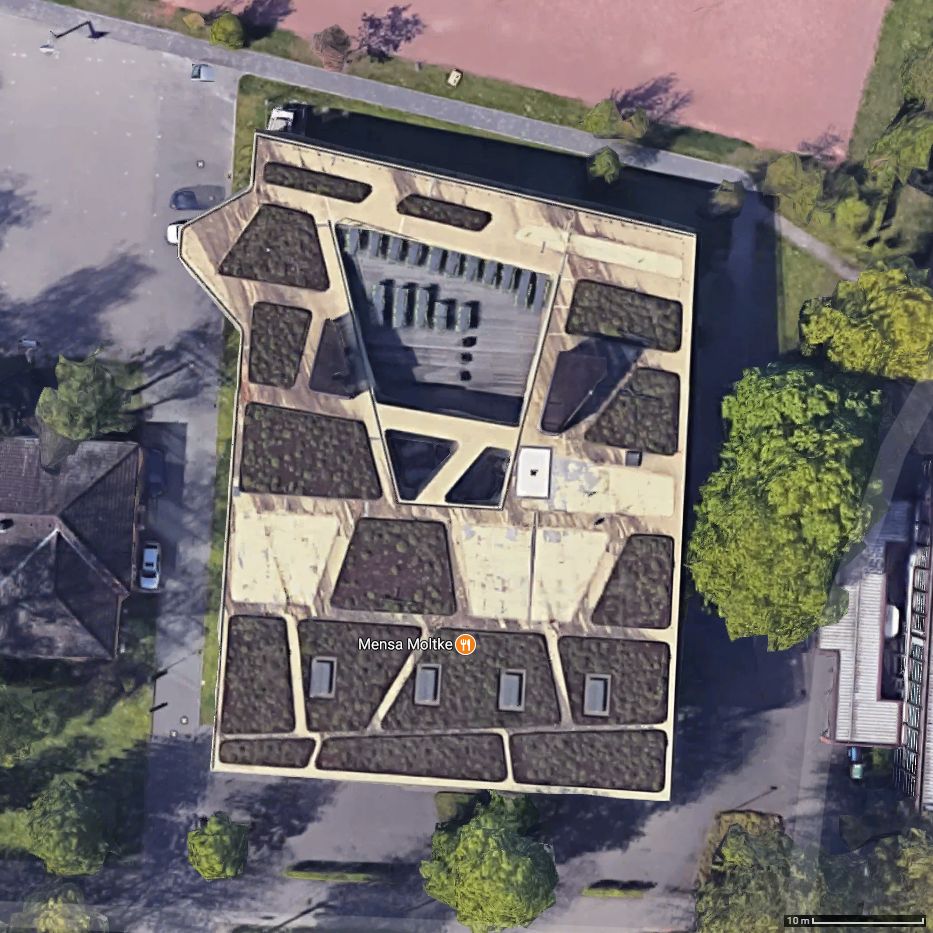}};
\node[inner sep=1pt, anchor=north west] at (image.north east){\rotatebox{-90}{\tiny source: GoogleMaps}};
\end{tikzpicture}}\hfill
\subfigure[estimated trajectory and 3D point cloud]{\includegraphics[width=0.48\linewidth]{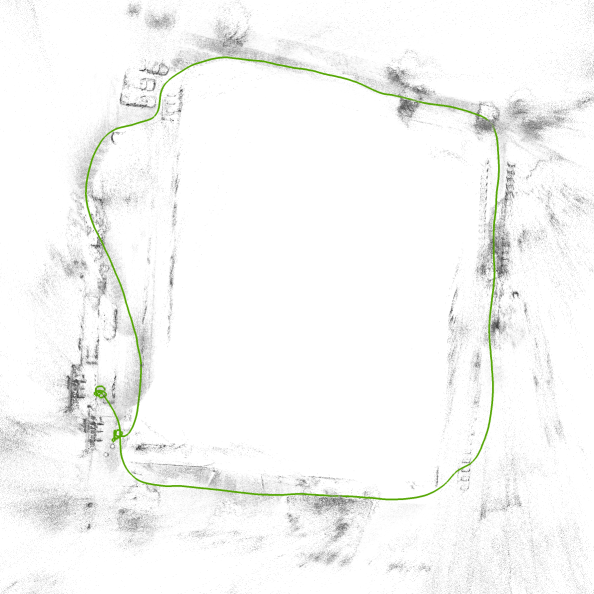}\label{fig:sample_complete_trajectory_2}}
\caption[Example sequence of the synchronized stereo and plenoptic \acs*{vo} dataset]{Example sequence of the dataset used for evaluation. (a) Top view of the real scene. (b) Trajectory (green) and \acs*{3d} point cloud estimated by \cite{Zeller2018a}.}
\label{fig:sample_complete_trajectory}
\end{figure}

To obtain reliable ground truth data, all sequences start and end in a scene showing objects in a distance of several meters, which are easy to track.
However, for these nearby objects it is easier to estimate the correct scale.
Hence, to consider only the scale drift $e_s'$ or the absolute scale $d_s$ might be misleading.
In combination with the alignment error  $e_\text{align}$, these values become more meaningful since the alignment error would reflect large scale drifts along the trajectory.

Following the scheme described above, we recorded a set of 11 sequences in versatile environments.
The recorded scenes range from large scales to small scales,
from man-made environments to environments with abundant vegetation.
The sequences capture moving objects like pedestrians, bikes or cars.
The sequences also cover difficult and changing lighting conditions due to shadows, moving clouds, and automatic exposure adjustment.
The path lengths of the performed trajectories range from \SI{25}{m} to \SI{274}{m}.

Figure~\ref{fig:sample_images_sequence} shows a set of sample images extracted from a single sequence.
The corresponding trajectory is shown in Figure~\ref{fig:sample_complete_trajectory}.
Figure~\ref{fig:point_cloud_gt_data} visualizes the registered start and end segments which are used to calculate the metrics.
As one can see, the monocular images of the stereo system (Fig.~\ref{fig:sample_images_sequence_stereo}) have a much wider \ac{fov} than the images of the plenoptic camera (Fig.~\ref{fig:sample_images_sequence_plenoptic}).
In Figure~\ref{fig:sample_images_sequence_plenoptic}, the images of the plenoptic camera seem to be a bit blurred.
This is not, in fact, the case and is only due to the multiple projections of a point in neighboring micro images.
Due to the narrower \ac{fov} and the higher number of pixels on the sensor, images of the plenoptic camera actually have a much higher spatial resolution than those from the monocular cameras.

\begin{figure}[t]
\centering
\includegraphics[width=\linewidth]{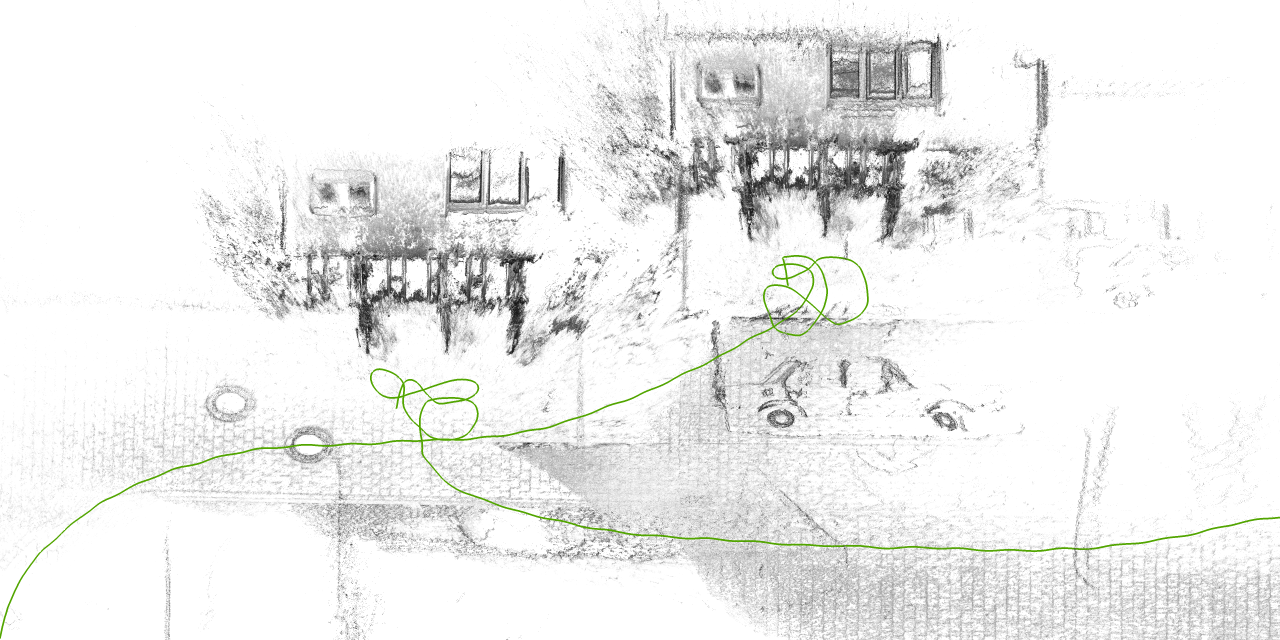}
\caption{Example of the accumulated drift of \cite{Zeller2018a} from the beginning to the end of a sequence. Due to the drift in the trajectory, the same scene is reconstructed twice at different locations.
The green line represents the camera trajectory estimated by the algorithm.}
\label{fig:point_cloud_drift}
\end{figure}

Figure~\ref{fig:point_cloud_drift} shows a magnified subsection of the trajectory and the point cloud  of Figure~\ref{fig:sample_complete_trajectory_2}.
This subsection shows the beginning and end of the sequence.
One can clearly see the drift accumulated over the complete sequence, resulting in the same scene being reconstructed twice in slightly different locations.

\section{Exemplary Results}\label{sec:results}

By way of example, this section shows results which were obtained for different algorithms based on the presented dataset.
The tested algorithms are:
\begin{itemize}
\item monocular:
	\begin{itemize}
	\item DSO \cite{Engel2017}
	\item ORB-SLAM2 \cite{Mur-Artal2015,Mur-Artal2017}
	\end{itemize}
\item stereo:
	\begin{itemize}
	\item ORB-SLAM2 \cite{Mur-Artal2017}
	\end{itemize}
\item plenoptic:
	\begin{itemize}
	\item SPO \cite{Zeller2018a}
	\end{itemize}
\end{itemize}
We also ran LSD-SLAM \cite{Engel2014} on the dataset.
Though, the algorithms failed on most of the sequences or resulted in extremely high drift metrics.

For none of the algorithms did we enforce real time processing.
For the algorithms which include a full \ac{slam} framework (ORB-SLAM2 and LSD-SLAM), large scale loop closure detection and relocalization was disabled.
The implementations of \acused{dso}\ac{dso} and LSD-SLAM are not able to handle the high image resolution of \SI{1.3}{mega pixel} of the monocular images.
Thus, for these algorithms, the image resolution is reduced to \SI{960 x 720}{pixel}.
Both versions of ORB-SLAM2 run at the full image resolution of \SI{1280 x 1024}{pixel}.

\begin{figure}
\centering
\subfigure[absolute scale error]{\begin{tikzpicture}
\begin{axis}[%
width=.8\linewidth,
height=.3\linewidth,
scale only axis,
xmin=1,
xmax=11,
xlabel={sequence number},
ymin=1,
ymax=1.4,
ymajorgrids,
ylabel={$d_s'$},
ybar,
enlarge x limits=0.07,
xtick align=inside,
xtick={1,...,11},
bar width=4,
legend style={at={(0.5,1.02)}, anchor=south,legend columns=-1,/tikz/every even column/.append style={column sep=0.5em}}
]
\addplot+[draw=gColor1,fill=gColor1] table [x expr={\thisrowno{0}},y expr={min(\thisrowno{2},1.4)},col sep=comma] {figures/results/results/abs_scale.csv};
\addlegendentry{SPO \cite{Zeller2018a}};
\addplot+[draw=gColor1,fill=gColor1!0] table [x expr={\thisrowno{0}},y expr={min(\thisrowno{1},1.4)},col sep=comma] {figures/results/results/abs_scale.csv};
\addlegendentry{SPO (no scale opt.)};
\addplot+[draw=gColor3,fill=gColor3!30] table [x expr={\thisrowno{0}},y expr={min(\thisrowno{7},1.4)},col sep=comma] {figures/results/results/abs_scale.csv};
\addlegendentry{ORB2 (stereo) \cite{Mur-Artal2017}};
\end{axis}
\end{tikzpicture}\label{fig:evaluation_abs_scale}}
\subfigure[scale drift]{\begin{tikzpicture}
\begin{axis}[%
width=.8\linewidth,
height=.3\linewidth,
scale only axis,
xmin=1,
xmax=11,
xlabel={sequence number},
ymin=1,
ymax=1.4,
ymajorgrids,
ylabel={$e_s'$},
ybar,
enlarge x limits=0.07,
xtick align=inside,
xtick={1,...,11},
bar width=4,
legend style={at={(0.5,1.02)}, anchor=south,legend columns=-1,/tikz/every even column/.append style={column sep=0.5em}}
]
\addplot+[draw=gColor1,fill=gColor1] table [x expr={\thisrowno{0}},y expr={min(\thisrowno{2},1.4)},col sep=comma] {figures/results/results/scale_drift.csv};
\addlegendentry{SPO \cite{Zeller2018a}};
\addplot+[draw=gColor2,fill=gColor2] table [x expr={\thisrowno{0}},y expr={min(\thisrowno{4},1.4)},col sep=comma] {figures/results/results/scale_drift.csv};
\addlegendentry{DSO \cite{Engel2017}}; 
\addplot+[draw=gColor3,fill=gColor3] table [x expr={\thisrowno{0}},y expr={min(\thisrowno{6},1.4)},col sep=comma] {figures/results/results/scale_drift.csv};
\addlegendentry{ORB2 (mono) \cite{Mur-Artal2017}};
\addplot+[draw=gColor3,fill=gColor3!30] table [x expr={\thisrowno{0}},y expr={min(\thisrowno{7},1.4)},col sep=comma] {figures/results/results/scale_drift.csv};
\addlegendentry{ORB2 (stereo) \cite{Mur-Artal2017}};
\end{axis}
\end{tikzpicture}\label{fig:evaluation_scale_drift}}
\subfigure[alignment error]{\begin{tikzpicture}
\begin{axis}[%
width=.8\linewidth,
height=.3\linewidth,
scale only axis,
xmin=1,
xmax=11,
xlabel={sequence number},
ymin=0,
ymax=6,
ymajorgrids,
ybar,
enlarge x limits=0.07,
ylabel={$e_\text{align}$ (in \%)},
xtick align=inside,
xtick={1,...,11},
bar width=4,
legend style={at={(0.5,1.02)}, anchor=south,legend columns=-1,/tikz/every even column/.append style={column sep=0.5em}}
]
\addplot+[draw=gColor1,fill=gColor1] table [x expr={\thisrowno{0}},y expr={min(\thisrowno{2}*100,6)},col sep=comma] {figures/results/results/alignment_relative.csv};
\addlegendentry{SPO \cite{Zeller2018a}};
\addplot+[draw=gColor2,fill=gColor2] table [x expr={\thisrowno{0}},y expr={min(\thisrowno{4}*100,6)},col sep=comma] {figures/results/results/alignment_relative.csv};
\addlegendentry{DSO \cite{Engel2017}}; 
\addplot+[draw=gColor3,fill=gColor3] table [x expr={\thisrowno{0}},y expr={min(\thisrowno{6}*100,6)},col sep=comma] {figures/results/results/alignment_relative.csv};
\addlegendentry{ORB2 (mono) \cite{Mur-Artal2017}};
\addplot+[draw=gColor3,fill=gColor3!30] table [x expr={\thisrowno{0}},y expr={min(\thisrowno{7}*100,6)},col sep=comma] {figures/results/results/alignment_relative.csv};
\addlegendentry{ORB2 (stereo) \cite{Mur-Artal2017}};
\end{axis}
\end{tikzpicture}\label{fig:alignment_error}}
\caption{Tracking drift measured based on the proposed dataset. (a) absolute scale errors, (b) scale drifts, and (c) alignment error for various monocular, stereo, and plenoptic \acl{vo} algorithms.
The alignment error is shown in percentages of the respective trajectory length.}
\label{fig:evaluation}
\end{figure}
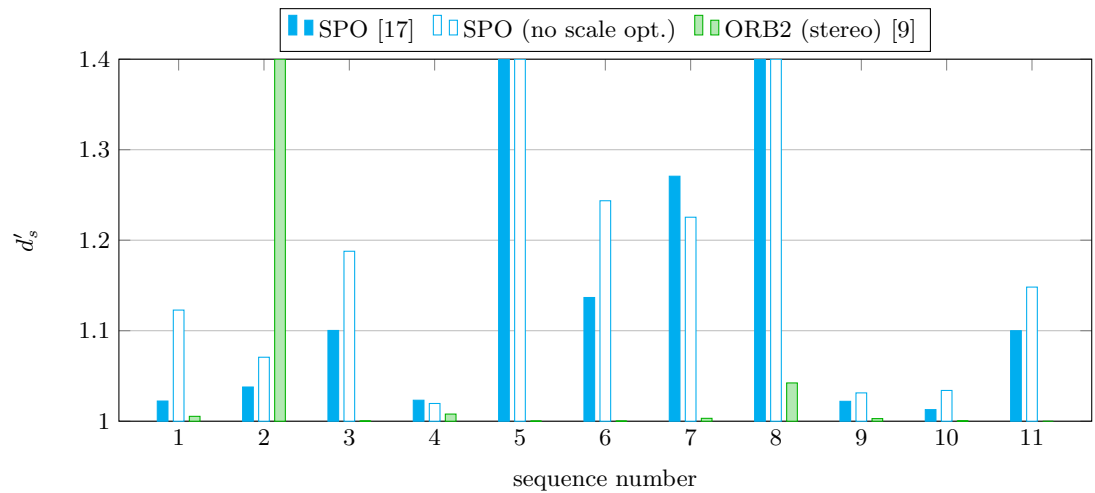
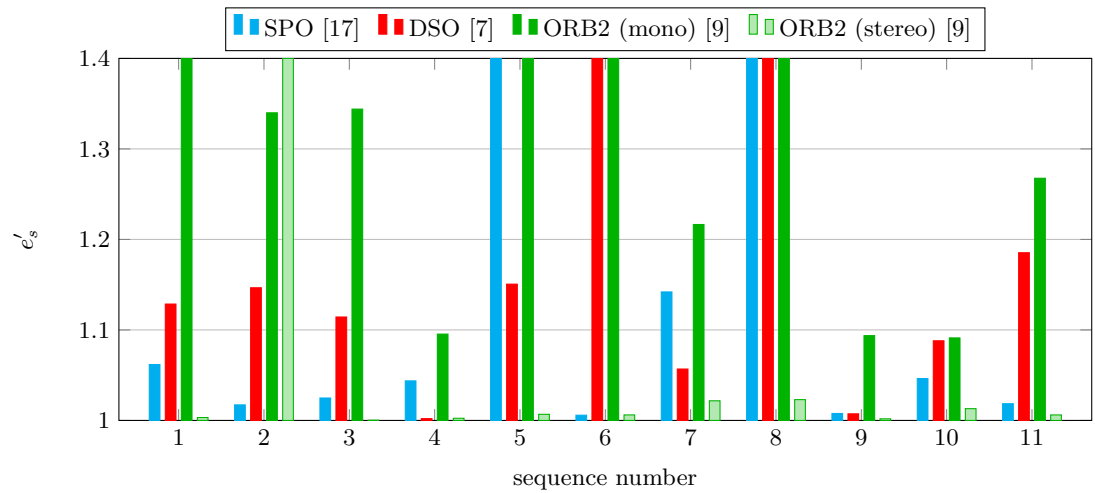
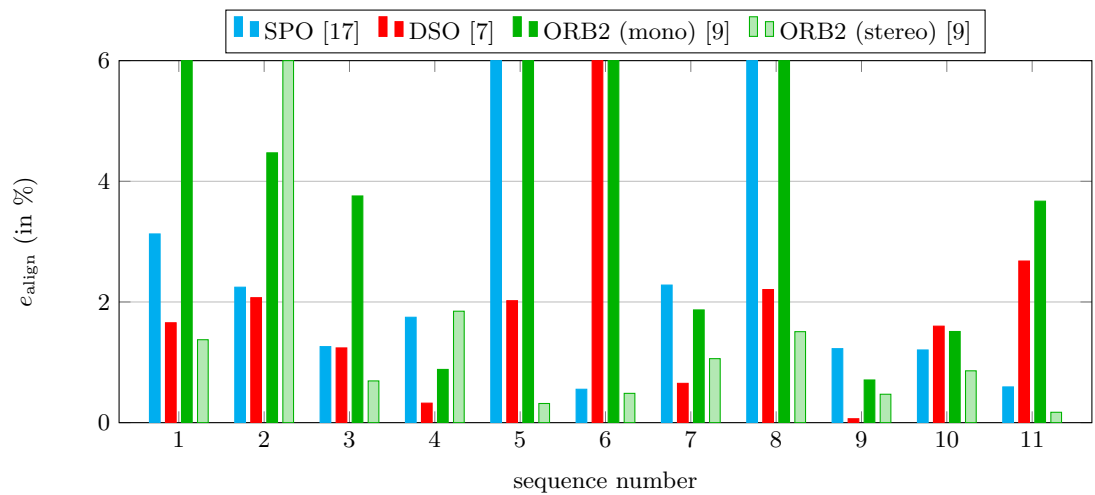

Figure~\ref{fig:evaluation} shows the results for all algorithms stated above, which were obtained based on the presented dataset.
Obviously, no absolute scale error $d_s'$ can be measured for the monocular algorithms.
Depending on the implementation a \ac{vo} algorithm either signals a tracking failure or results in an abnormally high tracking error.
In Figure~\ref{fig:evaluation} we chose appropriate graph limits.
All values at the upper graph border signify that the respective algorithm either failed, and therefore no metric could be measured, or that the measured metric lies above the upper graph limit.

\begin{figure}
\centering
\subfigure[]{\includegraphics[width=0.48\linewidth]{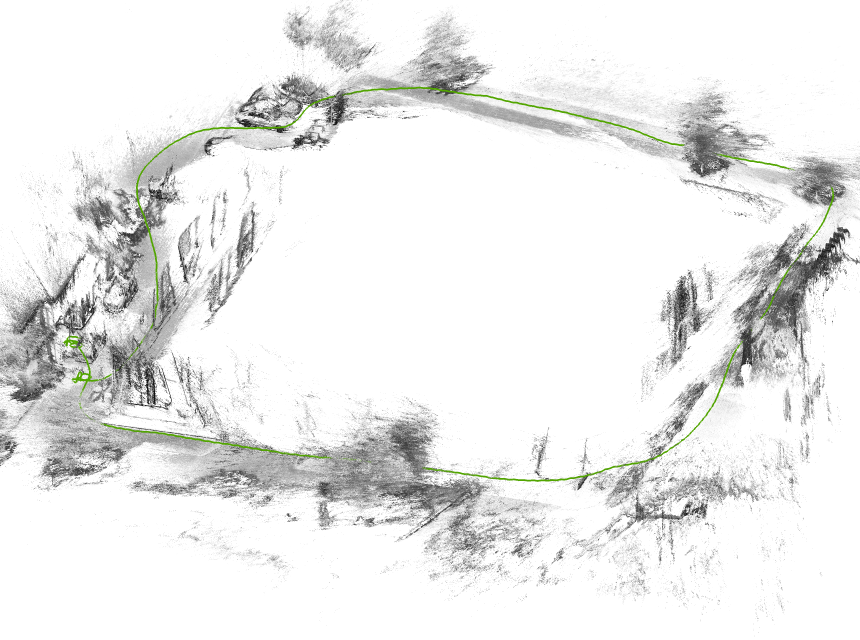}}\hfill
\subfigure[]{\includegraphics[width=0.48\linewidth]{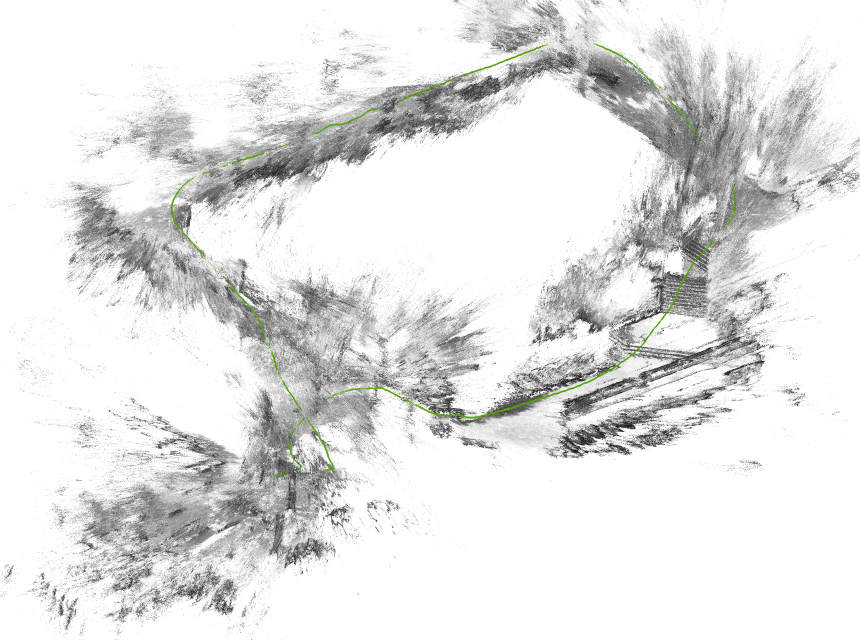}} 
\subfigure[]{\includegraphics[width=0.48\linewidth]{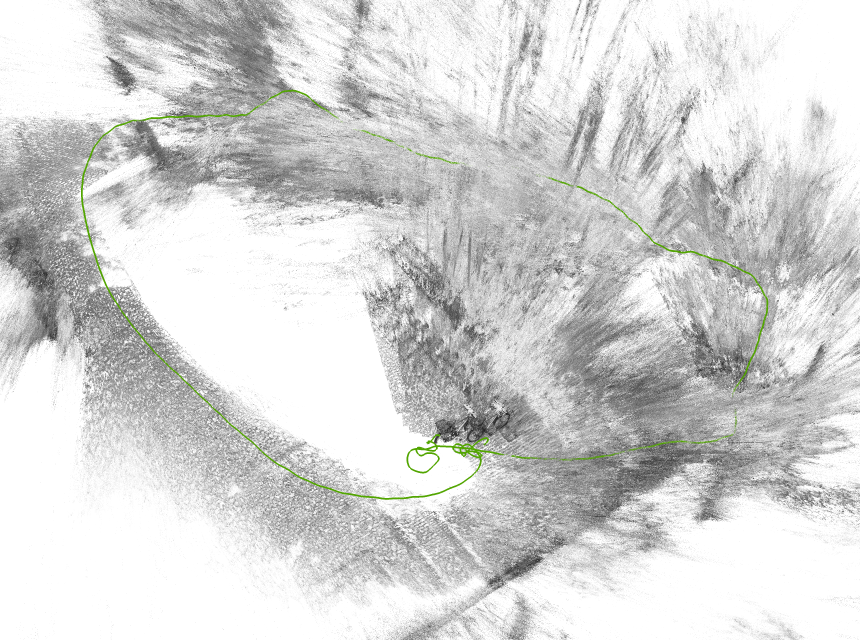}}\hfill
\subfigure[]{\includegraphics[width=0.48\linewidth]{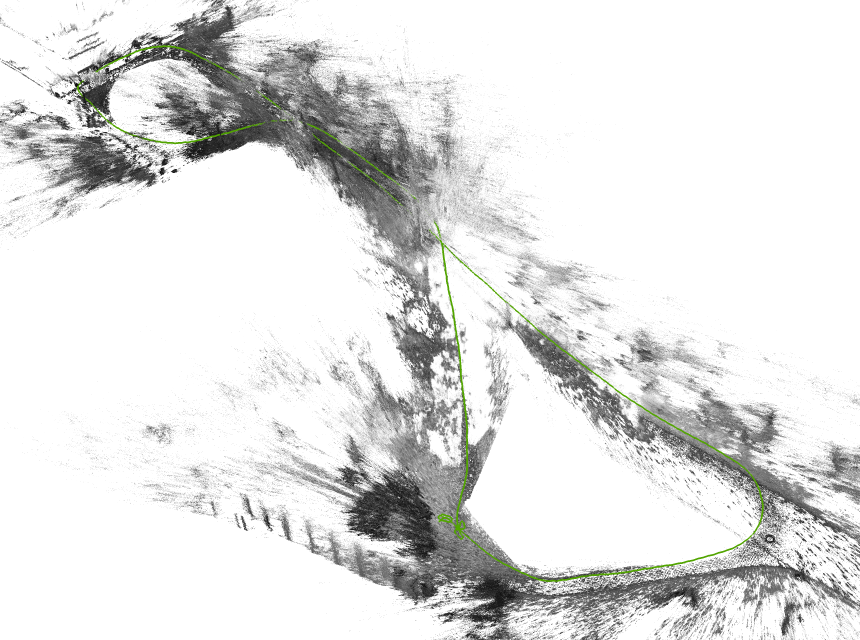}} 
\subfigure[]{\includegraphics[width=0.48\linewidth]{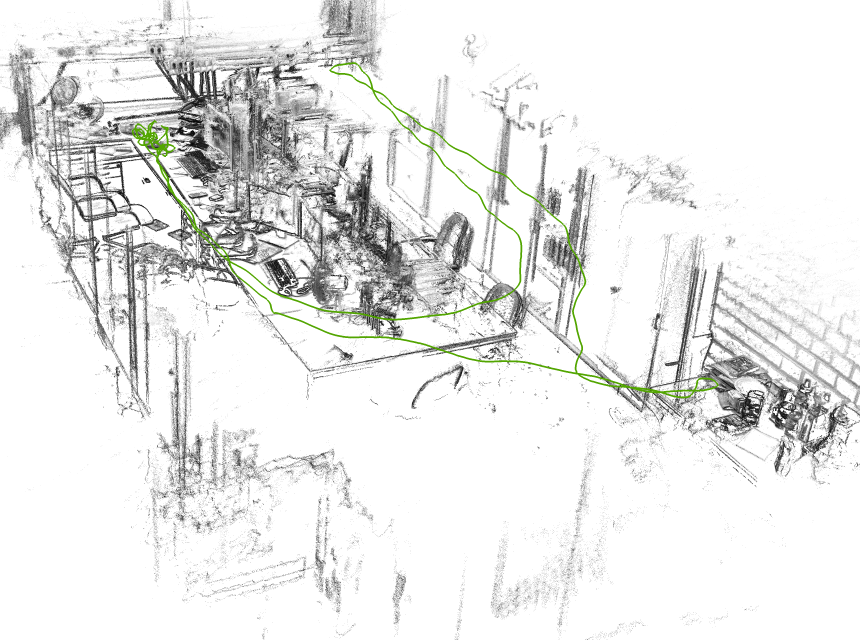}}\hfill
\subfigure[]{\includegraphics[width=0.48\linewidth]{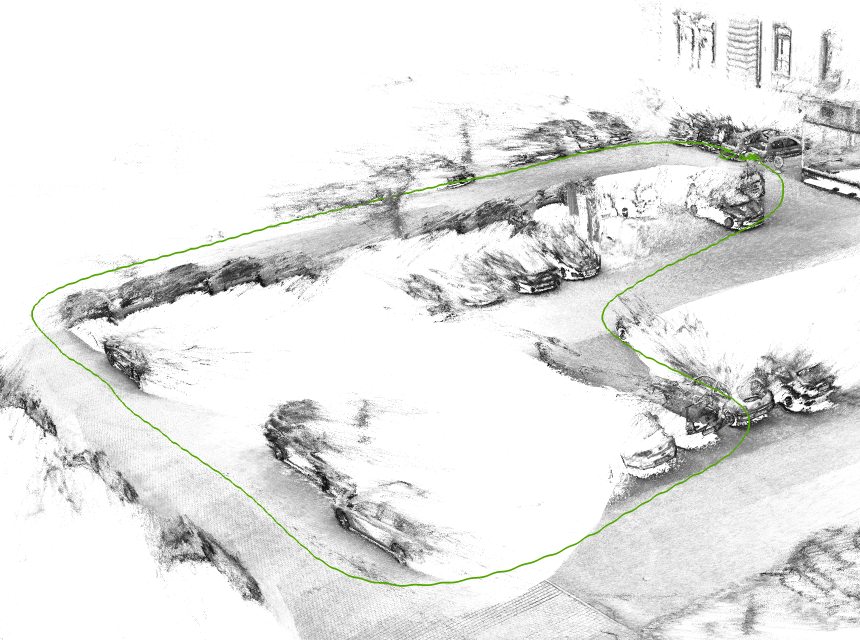}} 
\caption{Examples of point clouds reconstructed by \cite{Zeller2018a}.
The lengths of all 11 recorded trajectories range from \SI{25}{m} to \SI{274}{m}.}
\label{fig:example_point_clouds}
\end{figure}

Furthermore, Figure~\ref{fig:example_point_clouds} shows, by way of example, The 3D point clouds, for some of the sequences, calculated by \cite{Zeller2018a}.
These point clouds are supposed to give an impression about the recorded sequences.
More results can be found in \cite{Zeller2018a}.

From Figure~\ref{fig:evaluation} one can see that there are particular sequences for which the plenoptic camera based approach (SPO \cite{Zeller2018a}) perform worse than the other algorithms.
These sequences are on one side indoor sequences (e.g. \#7 and \#8) which
show corridors and staircases with lots of white walls. For these scenes monocular and stereo approaches benefit from the wider \acl{fov}.
On the other side SPO fails in an outdoor sequence (\#5), where a van is driving trough the scene.
Here, the monocular and stereo approaches again benefit from the wider \acl{fov}, while large areas of the light field image are covered by the driving car.

\section{Known Limitations}\label{sec:limitations}
While the two monocular cameras of the stereo system both have a monochromatic sensor, the plenoptic camera has a \acs{rgb} sensor.
Hence, even though all image sensors have a similar pixel size, a pixel of the plenoptic camera captures only approximately a third of the light energy compared to a pixel of the monocular cameras\footnote{For indoor sequences the light energy gathered by the plenoptic camera is even less than a third, since the ambient light here contains almost no infrared components and thus, red pixel are totally underexposed.}, when we, in fact, assume that all other parameters are similar.
For the plenoptic camera the F-number is predefined by construction, due to the aperture of the micro lenses.
This F-number is higher than the one of the lenses used for the two monocular cameras.
For this reason, the monocular cameras gather even more light energy on the same sensor area compared to the plenoptic camera.
To compensate for these two issues, a hardware-sided amplification of \SI{6}{dB} was set for the plenoptic camera.
Thus, for all cameras the exposure times are within the same order of magnitude, although they do not match exactly.
Due to the amplification, the images of plenoptic cameras will contain more noise when compared with the monocular cameras of the stereo system.

For the stereo camera system, it is important that both cameras run synchronized and with the same exposure time.
For this reason, the automatically calculated exposure time of the master camera must be used to set the exposure time of the slave camera.
It can happen that if the exposure time changes, an image pair is captured for which the two cameras had slightly different exposure times.

Currently, there exists no plenoptic camera based \ac{vo} algorithm which performs loop closures.
Therefore, the loop closure ground truth, calculated on the basis of the stereo images, is also used as ground truth for the plenoptic camera.
With respect to the plenoptic camera, the ground truth might be slightly inaccurate due to the slightly different positions of the master camera of the stereo system and the plenoptic camera.
The superior way would be to calculate a second ground truth on the basis of the plenoptic images.

Due to the reason that we use different sensors and lenses which have different properties, and the fact that the cameras see the scene from sightly different perspectives, one has to keep in mind that even though we are able to perform quantitative evaluations based on the presented dataset, these quantities are only valid up to a certain degree.
However, the dataset helps to emphasize the strength of \ac{vo} based on a certain sensor with respect to the other sensors.
The results presented in this paper as well as \cite{Zeller2018a} especially show that plenoptic camera based \ac{vo} offer a promising alternative to approaches based on traditional sensors.



\bibliographystyle{splncs}
\bibliography{vo_dataset_bib}

\begin{thebibliography}{10}

\bibitem{Klein2007}
Klein, G., Murray, D.:
\newblock Parallel tracking and mapping for small {AR} workspaces.
\newblock In: IEEE and ACM International Symposium on Mixed and Augmented
  Reality (ISMAR). Volume~6. (2007)  225--234

\bibitem{Newcombe2011}
Newcombe, R.A., Lovegrove, S.J., Davison, A.J.:
\newblock {DTAM}: Dense tracking and mapping in real-time.
\newblock In: IEEE International Conference on Computer Vision (ICCV). (2011)

\bibitem{Engel2013}
Engel, J., Sturm, J., Cremers, D.:
\newblock Semi-dense visual odometry for a monocular camera.
\newblock In: IEEE International Conference on Computer Vision (ICCV). (2013)
  1449--1456

\bibitem{Forster2014}
Forster, C., Pizzoli, M., Scaramuzza, D.:
\newblock {SVO}: Fast semi-direct monocular visual odometry.
\newblock In: IEEE International Conference on Robotics and Automation (ICRA).
  (2014)  15--22

\bibitem{Engel2014}
Engel, J., {Sch\"ops}, T., Cremers, D.:
\newblock {LSD-SLAM}: Large-scale direct monocular {SLAM}.
\newblock In: European Conference on Computer Vision (ECCV). (2014)  834--849

\bibitem{Mur-Artal2015}
Mur-Artal, R., Montiel, J.M.M., Tard\'{o}s, J.D.:
\newblock {ORB-SLAM}: A versatile and accurate monocular {SLAM} system.
\newblock IEEE Transactions on Robotics \textbf{31}(5) (2015)  1147--1163

\bibitem{Engel2017}
Engel, J., Koltun, V., Cremers, D.:
\newblock Direct sparse odometry.
\newblock IEEE Transactions on Pattern Analysis and Machine Intelligence
  \textbf{40}(3) (2018)  611--625

\bibitem{Engel2015}
Engel, J., St{\"u}ckler, J., Cremers, D.:
\newblock Large-scale direct {SLAM} with stereo cameras.
\newblock In: IEEE/RSJ International Conference on Intelligent Robots and
  Systems (IROS). (2015)  1935--1942

\bibitem{Mur-Artal2017}
Mur-Artal, R., Tard\'{o}s, J.D.:
\newblock {ORB-SLAM2}: An open-source {SLAM} system for monocular, stereo, and
  {RGB-D} cameras.
\newblock IEEE Transactions on Robotics \textbf{33}(5) (2017)  1255--1262

\bibitem{Wang2017}
Wang, R., Schw\"orer, M., Cremers, D.:
\newblock Stereo {DSO}: Large-scale direct sparse visual odometry with stereo
  cameras.
\newblock In: International Conference on Computer Vision (ICCV). (2017)

\bibitem{Izadi2011}
Izadi, S., Kim, D., Hilliges, O., Molyneaux, D., Newcombe, R., Kohli, P.,
  Shotton, J., Hodges, S., Freeman, D., Davison, A., Fitzgibbon, A.:
\newblock {KinectFusion}: Real-time {3D} reconstruction and interaction using a
  moving depth camera.
\newblock In: 24th Annual ACM Symposium on User Interface Software and
  Technology, ACM (2011)  559--568

\bibitem{Kerl2013}
Kerl, C., Sturm, J., Cremers, D.:
\newblock Dense visual {SLAM} for {RGB-D} cameras.
\newblock In: IEEE/RSJ International Conference on Intelligent Robots and
  Systems (IROS). (2013)  2100--2106

\bibitem{Kerl2015}
Kerl, C., St{\"u}ckler, J., Cremers, D.:
\newblock Dense continuous-time tracking and mapping with rolling shutter
  {RGB-D} cameras.
\newblock In: IEEE International Conference on Computer Vision (ICCV). (2015)
  2264--2272

\bibitem{Dansereau2011}
Dansereau, D., Mahon, I., Pizarro, O., Williams, S.:
\newblock Plenoptic flow: Closed-form visual odometry for light field cameras.
\newblock In: IEEE/RSJ International Conference on Intelligent Robots and
  Systems (IROS). (2011)  4455--4462

\bibitem{Dong2013}
Dong, F., Ieng, S.H., Savatier, X., Etienne-Cummings, R., Benosman, R.:
\newblock Plenoptic cameras in real-time robotics.
\newblock The International Journal of Robotics Research \textbf{32}(2) (2013)
  206--217

\bibitem{Zeller2017}
Zeller, N., Quint, F., Stilla, U.:
\newblock From the calibration of a light-field camera to direct plenoptic
  odometry.
\newblock IEEE Journal of Selected Topics in Signal Processing \textbf{11}(7)
  (2017)  1004--1019

\bibitem{Zeller2018a}
Zeller, N., Quint, F., Stilla, U.:
\newblock Scale-awareness of light field camera based visual odometry.
\newblock In: European Conference on Computer Vision (ECCV). (2018)

\bibitem{Zhang2016b}
Zhang, C., Rebecq, H., Forster, C., Scaramuzza:
\newblock Benefit of large field-of-view cameras for visual odometry.
\newblock In: IEEE International Conference on Robotics and Automation (ICRA).
  (2016)  801--808

\bibitem{Engel2016a}
Engel, J., Usenko, V., Cremers, D.:
\newblock A photometrically calibrated benchmark for monocular visual odometry.
\newblock In: arXiv:1607.02555. (2016)

\bibitem{Majdik2017}
Majdik, A.L., Till, C., Scaramuzza, D.:
\newblock The {Zurich} urban micro aerial vehicle dataset.
\newblock The International Journal of Robotics Research \textbf{36}(3) (2017)
  269--273

\bibitem{Geiger2012}
Geiger, A., Lenz, P., Urtasun:
\newblock Are we ready for autonomous driving? the {KITTI} vision benchmark
  suite.
\newblock In: IEEE Conference on Computer Vision and Pattern Recognition
  (CVPR). (2012)  3354--3361

\bibitem{Burri2016}
Burri, M., Nikolic, J., Gohl, P., Schneider, T., Rehder, J., Omari, S.,
  Achtelik, M.W., Siegwart, R.:
\newblock The {EuRoC} micro aerial vehicle datasets.
\newblock Internation Journal of Robotics Research \textbf{35}(10) (2016)
  1157--1163

\bibitem{Schoeps2017}
Sch{\"o}ps, T., Sch{\"o}nberger, J.L., Galliani, S., Sattler, T., Schindler,
  K., Pollefeys, M., Geiger, A.:
\newblock A multi-view stereo benchmark with high-resolution images and
  multi-camera videos.
\newblock In: IEEE Conference on Computer Vision and Pattern Recognition
  (CVPR). (2017)

\bibitem{Sturm2012}
Sturm, J., Engelhard, N., Endres, F., Burgard, W., Cremers:
\newblock A benchmark for the evaluation of {RGB-D} slam systems.
\newblock In: IEEE/RSJ International Conference on Intelligent Robot Systems
  (IROS). (2012)

\bibitem{Handa2014}
Handa, A., Whelan, T., McDonald, J., Davison:
\newblock A benchmark for {RGB-D} visual odometry, {3D} reconstruction and
  {SLAM}.
\newblock In: IEEE International Conference on Robotics and Automation (ICRA).
  (2014)  1524--1531

\bibitem{Dansereau2013}
Dansereau, D., Pizarro, O., Williams, S.:
\newblock Decoding, calibration and rectification for lenselet-based plenoptic
  cameras.
\newblock In: IEEE Conference on Computer Vision and Pattern Recognition
  (CVPR). (2013)  1027--1034

\end{thebibliography}

\end{document}